%% file: acl_latex.tex
\documentclass[11pt]{article}

\usepackage[preprint]{acl}

\usepackage{times}
\usepackage{latexsym}

\usepackage[T1]{fontenc}
\usepackage[utf8]{inputenc}
\usepackage{microtype}
\usepackage{inconsolata}

\usepackage{graphicx}
\usepackage{subcaption}
\usepackage{booktabs}
\usepackage{arydshln}
\usepackage{pdfpages}
\usepackage{bm}
\usepackage{multirow}
\usepackage{enumitem}
\usepackage[table]{xcolor}

\usepackage{amsmath}
\usepackage{amssymb}
\usepackage{mathtools}
\usepackage{amsthm}

\theoremstyle{plain}

\theoremstyle{definition}

\theoremstyle{remark}

\usepackage[textsize=tiny]{todonotes}

\title{Unified Data Selection for LLM Reasoning}

\author{
  \textbf{Xiaoyuan Li\textsuperscript{1}},
  \textbf{Yubo Ma\textsuperscript{2}},
  \textbf{Chengpeng Li\textsuperscript{2}},
  \textbf{Fengbin Zhu\textsuperscript{3}},
  \textbf{Yiyao Yu\textsuperscript{2}},
\\
  \textbf{Keqin Bao\textsuperscript{2}},
  \textbf{Wenjie Wang\textsuperscript{1}},
  \textbf{Fuli Feng\textsuperscript{1}},
  \textbf{Dayiheng Liu\textsuperscript{2}}
\\
\\
  \textsuperscript{1}University of Science and Technology of China,
  \textsuperscript{2}Alibaba Group,
  \\
  \textsuperscript{3}National University of Singapore
}

\begin{document}
\maketitle

\input{section/1_abstract}
\input{section/2_introduction_v5}

\input{section/3_preliminaries}
\input{section/4_method_0128}

\input{section/5_experiment_0128}
\input{section/6_related_work}
\input{section/7_conclusion}

\bibliography{custom}

\appendix
\input{section/appendix}

\end{document}

%% file: section/1_abstract.tex
\begin{abstract}
Effectively training Large Language Models (LLMs) for complex, long-CoT reasoning is often bottlenecked by the need for massive high-quality reasoning data. Existing methods are either computationally expensive or fail to reliably distinguish high- from low-quality reasoning samples. 
To address this, we propose \textbf{High-Entropy Sum (HES)}, a training-free metric that quantifies reasoning quality by summing only the entropy of the top (\emph{e.g.}, 0.5\%) highest-entropy tokens in each reasoning sample.
We validate HES across three mainstream training paradigms:  Supervised Fine-
tuning (SFT), Rejection Fine-tuning
(RFT), and Reinforcement Learning
(RL), with extensive results demonstrating its consistent effectiveness and significantly reduced computational overhead. 
In SFT, training on the top 20\% HES-ranked data matches full-dataset performance, while using the lowest-HES data degrades it. In RFT, our HES-based training approach significantly outperforms baseline methods. In RL, HES-selected successful trajectories enable the model to learn strong reasoning patterns, significantly surpassing other compared methods.
Our findings establish HES as a robust, training-free metric that enables a unified, effective, and efficient method for developing advanced reasoning in LLMs.
\end{abstract}


%% file: section/2_introduction_v5.tex
\section{Introduction}
The ability of Large Language Models (LLMs) to solve complex problems through Chain-of-Thoughts (CoT) reasoning has become a central focus in frontier research~\citep{jaech2024openai,deepseekai2025,yang2025qwen3}. To enhance models' reasoning capabilities, dominant training paradigms, including Supervised Fine-tuning (SFT)~\citep{ouyang2022training}, Rejection Fine-tuning (RFT)~\citep{yuan2023scaling}, and Reinforcement Learning (RL)~\citep{shao2024deepseekmath}, heavily rely on high-quality training data. However, the indiscriminate expansion of training data often introduces more noise and incurs additional costs, spurring a critical need for efficient and effective data selection methods~\citep{zhou2023lima}.
The key lies in defining a robust metric to rapidly and accurately distinguish high-quality data from low-quality data~\citep{li2024quantity}.

\input{figure/aime_2025}

Current research has explored various strategies for filtering training data, such as filtering by length ~\citep{rae2021scaling}, perplexity ~\citep{marion2023less}, or average token entropy \citep{sabbineni2023comprehensive}.
However, these metrics share common limitations when applied to long-CoT reasoning scenarios, where extended and methodical reasoning processes are particularly valuable for model learning ~\citep{chen2025towards}. 
Specifically, they rely on a \textbf{coarse-grained, global evaluation} of the reasoning path, assigning equal weight to all tokens. 
This uniform averaging fails to capture the nuanced, multi-stage structure of complex reasoning, which often involves planning, exploration, and reflection \citep{chen2025towards}. 
Although some methods address this by training auxiliary task-specific models for data selection ~\citep{shum2025predictive} or by employing powerful LLMs to choose among multiple responses~\citep{toshniwal2025genselect}, such approaches incur \textbf{substantial computational overhead}. 
Moreover, because these methods are not directly aligned with or adaptive to the target model being optimized, they risk yielding suboptimal training data.

To address the above gaps, our objective is to develop an \textbf{efficient} method for selecting high-quality reasoning samples by identifying and leveraging \textbf{fine-grained, critical information} embedded in long-CoT responses.
Recently, \cite{wang2025beyond} has suggested that a small number of high-entropy \textit{critical tokens} in the reasoning process serve as a key driver for performance improvement.
Inspired by this, we conduct a preliminary study on token-level entropy distribution throughout the reasoning process.
Note that high-entropy tokens are identified as those falling within the top 0.5\% of the entropy distribution across all tokens.
We present the results in Figure \ref{aime_2025}, from which we can observe: 1) Neither the overall average token entropy (Fig. \ref{aime_2025} (a) ) nor the average entropy of high-entropy tokens (Fig. \ref{aime_2025} (b)) can differentiate high- from low-quality samples.
2) Comparably, the summation of entropy over high-entropy tokens (Fig. \ref{aime_2025} (d)) provides a more robust distinction compared to the total entropy of all tokens (Fig. \ref{aime_2025} (c)).

These findings demonstrate a strong correlation between sample quality and the cumulative entropy of \emph{critical tokens}.
Capitalizing on this relationship, we propose \textbf{High-Entropy Sum (HES)}, a train-free metric to quantify the long-CoT reasoning quality of the samples.
Specifically, HES is calculated by summing the entropy values of the top  $k\%$ (\textit{e.g.}, 0.5\%) of tokens with the highest entropy.
Building on this metric, we propose an efficient, unified data selection method to curate high-quality reasoning data for LLMs, applicable across SFT, RFT, and RL training paradigms.

To evaluate the effectiveness of our proposed HES-based method, we conduct extensive experiments across the three training paradigms. 
The results reveal that:
1) \textbf{In SFT}, training on the top 20\% of samples ranked by HES achieves performance comparable to the full dataset across diverse domains, while expanding to the top 80\% consistently surpasses the baseline. Notably, we demonstrate that data selection using a lightweight 0.6B model yields results comparable to those of an 8B model, validating the cost-effectiveness of HES for offline data curation; 2) \textbf{In RFT}, HES significantly outperforms traditional random selection, serving as an effective, training-free reward signal; 3) \textbf{In RL}, our strategy of oversampling rollouts and selecting only the top-50\% ranked by HES outperforms the standard setting where all rollouts participate in policy updates. Collectively, these findings highlight that HES is not only high-performing but also remarkably efficient: it enables scalable proxy selection with small models in offline SFT and incurs negligible storage overhead in online RFT and RL settings.
The empirical results confirm that our HES-based method is both an effective and efficient approach for selecting high-quality data, thereby advancing the reasoning capabilities of LLMs.

In summary, our contributions are as follows:
\begin{itemize}[leftmargin=*]
\item We introduce High-Entropy Sum (HES), a train-free metric that quantifies reasoning quality by analyzing only the high-entropy tokens in the response paths.
\item We propose an effective and efficient HES-based data selection method that is compatible with SFT, RFT, and RL training paradigms.
\item Extensive experiments across diverse domains demonstrate that HES consistently outperforms baselines in SFT, RFT, and RL, achieving superior model performance with significantly reduced computational overhead.

\end{itemize}

%% file: figure/aime_2025.tex
\begin{figure*}[t]   
\centering
\setlength{\abovecaptionskip}{-0.05cm}
\includegraphics[width=1.0\linewidth,scale=1.0]{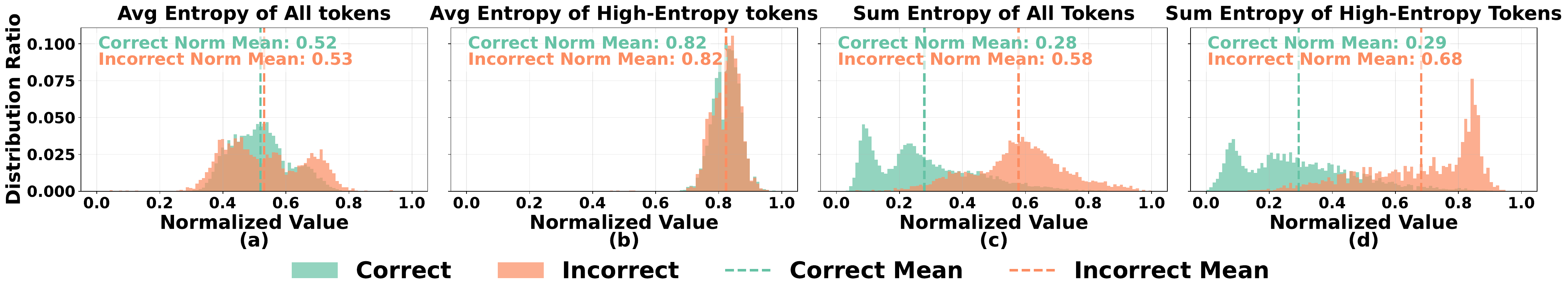}
\caption{The comparative analysis of discriminative ability is conducted on 512 responses per problem generated by Qwen3-14B for AIME 2025. The results demonstrate that metric of \emph{(d) Sum Entropy of High-Entropy Tokens} is the most effective to distinguish between high- and low-quality samples.}
\label{aime_2025}
\vspace{-10pt}
\end{figure*}

%% file: section/3_preliminaries.tex
\vspace{-10pt}
\section{Preliminaries}\label{global}

\input{figure/example}

\subsection{Training Paradigms}

\paragraph{Supervised Fine-Tuning (SFT).}
Let $\mathcal{D}=\{(x,y)\}$ denote a corpus of correct demonstrations, where $x$ is a query and $y$ is the reference response. SFT minimizes the cross-entropy loss, defined by the objective function:
$\mathcal{L}_{SFT}(\theta) = \mathbb{E}_{(x,y)\sim\mathcal{D}}[-\log \pi_{\theta}(y|x)]$,
where $\theta$ represents the model's parameters and $\pi_{\theta}(y|x)$ is the probability assigned by the model to $y$ given $x$.

\paragraph{Rejection Sampling Fine-Tuning (RFT).} RFT~\citep{yuan2023scaling} augments SFT by generating training samples through the model's own exploration. The process involves three steps:
(1) Generation: For query $x$, generate $m$ different candidate responses $\{y_1, y_2, \dots, y_m\}$;
(2) Selection: Use selection function $R(y)$ to select a response subset $Y$:
$Y = \underset{y_i \in \{y_1, \dots, y_m\}}{\text{argmax}} \, R(y_i)$;
(3) Fine-tuning: Create new dataset $\mathcal{D}^{*}$ with $Y$ and fine-tune using SFT.

\paragraph{Reinforcement Learning (RL).} Group Relative Policy Optimization (GRPO)~\citep{shao2024deepseekmath} is a variant of Proximal Policy Optimization (PPO)~\citep{schulman2017proximal}. For each query, GRPO samples a group of $G$ responses $\{o_1, \dots, o_G\}$ with corresponding rewards $\{r_1, \dots, r_G\}$. The advantage $\hat{A}_i$ for response $o_i$ is first computed as:
$\hat{A}_{i} = \frac{r_i - \text{mean}(\{r_j\}_{j=1}^{G})}{\text{std}(\{r_j\}_{j=1}^{G})} $.
This group-normalized advantage is then used in a clipped policy gradient objective to update parameters $\theta$. The full objective function to be maximized is:
$\mathcal{J}_{\text{GRPO}}(\theta) = $  
$\mathbb{E} [ \min\left( r_t(\theta) \hat{A}_{i}, \text{clip}(r_t(\theta), 1 - \epsilon, 1 + \epsilon) \hat{A}_{i} \right) - $
$\beta D_{\text{KL}}(\pi_{\theta_{\text{old}}} \| \pi_{\theta})]$.
Here, $r_t(\theta)$ is the probability ratio, $\hat{A}_{i}$ is the group-relative advantage, $\epsilon$ is a clipping bound, and the final term is a KL-divergence penalty with weight $\beta$ to stabilize training.

\subsection{Token Entropy}

\paragraph{Token entropy.} For a token position $t$ with probability distribution $P_t$ over the vocabulary, the token entropy $H_t$ is given by: $H_t = -\sum_{j} P_t(j) \log P_t(j)$, where $P_t(j)$ represents the predicted probability for the $j$-th token. Token entropy is the fundamental metric to measure a model's uncertainty during generation. Low entropy typically occurs during predictable parts of a reasoning path, such as completing a common phrase, performing a simple calculation, or following a standard template. Conversely, high entropy signifies high uncertainty where the model is considering multiple viable and often competing options. In the context of long-CoT reasoning, these high-entropy moments are particularly important as they often correspond to critical forks where the model must make a non-trivial decision that will shape the subsequent trajectory~\citep{wang2025beyond}.

\vspace{-10pt}
\paragraph{Average Entropy.} The common approach to quantifying the overall uncertainty of a reasoning path is average entropy~\citep{sabbineni2023comprehensive}, defined for a path of length $N$ as: $\text{AvgE} = \frac{1}{N} \sum_{t=1}^{N} H_t = \frac{1}{N} \text{ES}$, where $\text{ES}$ is the entropy sum of all tokens. However, average entropy is limited because it masks critical local signals by averaging over long sequences. A high-quality reasoning path that successfully navigates multiple challenging forks may receive a similar score to one that follows a straightforward, low-complexity approach. This inability to distinguish between different reasoning complexity makes the metric unreliable for identifying the most valuable samples.

%% file: figure/example.tex
\begin{figure*}[t]   
\centering
\setlength{\abovecaptionskip}{-0.05cm}
\includegraphics[width=1.0\linewidth,scale=1.0]{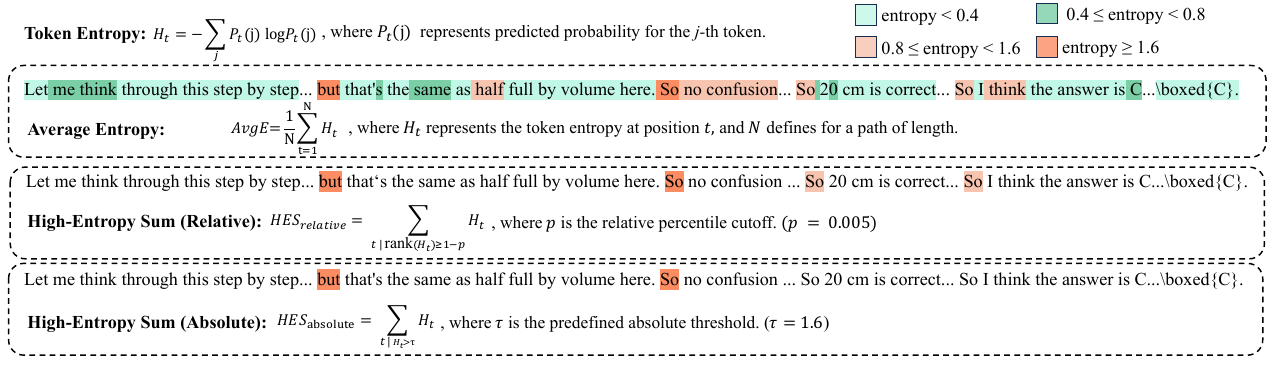}
\caption{Calculation of High-Entropy Sum (HES). The colored tokens are the tokens involved in the calculation.}
\label{example}
\vspace{-10pt}
\end{figure*}

%% file: section/4_method_0128.tex
\section{Methodology}
\input{table_0128/rl_method}
To overcome the limitations of global metrics, we first introduce High-Entropy Sum (HES) as a quantitative measure of reasoning quality and then leverage it to guide data selection across three major training paradigms. 

\subsection{High-Entropy Sum}
Our core metric is HES, which quantifies the cumulative intensity of pivotal, high-entropy moments in a reasoning path. We define this metric in two primary formulations as shown in Figure~\ref{example}: a robust, adaptive version using a relative, percentile-based threshold ($\text{HES}_{\textit{relative}}$), and a simpler alternative using a fixed, absolute threshold ($\text{HES}_{\textit{absolute}}$).

\paragraph{$\text{HES}_{\textit{relative}}$.} This metric is designed to capture the cumulative complexity of a sample by focusing on its most uncertain tokens in an adaptive manner.

{
\small
\begin{equation}
\text{HES}_{\textit{relative}}= \sum_{t \, | \, \text{rank}(H_t) \ge 1-p} H_t.
\end{equation}
}

Here, $\text{rank}(H_t)$ selects the tokens $t$ whose entropy $H_t$ rank within the top $p$-th percentile of a sample for summation (e.g., $p=0.005$ for the top 0.5\%). A higher $\text{HES}_{\textit{relative}}$ score indicates the successful navigation of more numerous and intense forks, which represents higher quality. The relative threshold makes this metric robust to variations in length among different models and tasks.\footnote{As experiments will demonstrate, $\text{HES}_{\textit{relative}}$ provides better results than $\text{HES}_{\textit{absolute}}$. This is because its adaptive nature makes it more robust across the diverse entropy distributions found in different models and reasoning paths. So, for the remainder, HES will refer to $\text{HES}_{\textit{relative}}$ unless otherwise specified.}

\paragraph{$\text{HES}_{\textit{absolute}}$.} Alternatively, this metric is used for quality estimation by a fixed cutoff.

{
\small
\begin{equation}
    \text{HES}_{\textit{absolute}} = \sum_{t \, | \, H_t > \tau} H_t.
\end{equation}
}

Here, the sum is taken over the tokens $t$ whose entropy $H_t$ exceeds a predefined absolute value $\tau$ (e.g., $\tau=1.6$). While less adaptive than the relative threshold, this method provides a straightforward alternative for quality estimation, particularly in contexts where a consistent entropy scale can be expected across all samples.

\paragraph{\textbf{$\text{AvgHE}$}.}
To test whether the cumulative sum of uncertainty is a more effective signal than its average intensity, we introduce average entropy of high-entropy tokens.

{
\small
\begin{equation}
    \text{AvgHE}= \frac{1}{|T_{high}|} \sum_{t \in T_{high}} H_t = \frac{\text{HES}}{|T_{high}|}.
\end{equation}
}

For this metric, we identify the set of high-entropy tokens $T_{high}$ using the relative threshold described above. $\text{AvgHE}$ is defined as $\text{HES}_{\textit{relative}}$ normalized by $|T_{high}|$, the number of tokens in the set. It is designed to isolate the average complexity of key-fork tokens, different from $\text{HES}$.

\subsection{Unified HES-Guided Data Selection}
\label{sec:method_selection}

We apply the proposed HES metric to curate data and guide training across three major paradigms: SFT, RFT, and RL. 

\paragraph{Preliminaries.} Let $\mathcal{D} = \{(x, y)\}$ denote a dataset where $x$ represents the prompt and $y$ represents the reasoning response. Let $S(y)$ denote the HES score of a response $y$, calculated as defined in Eq. (1). We define a selection ratio $\rho \in (0, 1]$ to control the proportion of data retained.

\paragraph{Selection for SFT.}
In the SFT stage, our goal is to identify a high-quality subset $\mathcal{D}_{\text{SFT}}$ from the full dataset $\mathcal{D}$. We rank all samples based on their HES scores and select the top-$\rho$ percentile. Formally, the curated dataset is defined as:
\begin{equation}
    \mathcal{D}_{\text{SFT}}(\rho) = \{ (x, y) \in \mathcal{D} \mid S(y) \geq \tau_\rho \}
\end{equation}
where $\tau_\rho$ is the score threshold corresponding to the $(1-\rho)$-th percentile of the HES distribution. 
We primarily investigate two settings: (1) \textbf{Efficiency Test} ($\rho=0.2$), training on the top 20\% highest-HES samples to evaluate sample efficiency; and (2) \textbf{De-noising Test} ($\rho=0.8$), pruning the bottom 20\% lowest-HES samples to test if removing low-quality data improves performance over the full dataset.

\paragraph{Selection for RFT.}
For RFT, we first generate $K$ candidate responses $\mathcal{Y}_i = \{y_{i,1}, \dots, y_{i,K}\}$ for each query $x_i$, and filter them to retain only the correct solutions, denoted as $\mathcal{Y}_i^+$. We then apply HES selection in two scopes:

\begin{itemize}[leftmargin=*]
    \item \textbf{Per-Query Selection:} For each query $x_i$, we select the top-$k$ responses with the highest HES scores from the candidate set: $\mathcal{Y}_i^* = \text{Top-}k(\{y \in \mathcal{Y}_i^+ \mid S(y)\})$. This tests HES's ability to identify high-quality solutions from a localized pool with a fixed budget per problem\footnote{If the number of $\mathcal{Y}_i^+$ is less than $k$, then all of $\mathcal{Y}_i^+$ are taken. In our experiments, $K$ is 32 and $k=2,4,8$.}.
    
    \item \textbf{Global Pool Selection:} We aggregate all correct responses from all queries into a global pool $\mathcal{P}_{\text{global}} = \bigcup_i \mathcal{Y}_i^+$. We then perform selection on this global set, retaining the top-$N$ samples with the highest HES scores (where $N$ matches the total data volume of the per-query setting). This challenging setting tests whether HES can distinguish high-quality reasoning across a diverse, global distribution.
\end{itemize}

\paragraph{Selection for RL.}
In the RL stage, the model generates a group of outputs for each prompt. Let $Y^+$ and $Y^-$ denote the sets of correct and incorrect trajectories, respectively. Unlike standard approaches that sample uniformly or utilize the full batch, we propose an \textit{asymmetric sampling strategy} that differentiates the selection logic for positive and negative signals to maximize training efficiency.

Our primary strategy, \textbf{Pos-High, Neg-Rand}, prioritizes the highest-HES positive samples to encourage the mastery of complex reasoning patterns, while retaining random negative samples to ensure robust exposure to diverse error modes. To systematically isolate the contributions of these components, we define a comprehensive sampling space encompassing five distinct strategies, as detailed in Table~\ref{tab:rl_sampling_strategies}. This framework allows us to validate the efficacy of HES in identifying valuable positive reinforcement signals while demonstrating the necessity of unbiased negative feedback.

%% file: table_0128/rl_method.tex
\begin{table*}[t]
    \centering
    \caption{Definition of asymmetric sampling strategies for RL. $Y^+$ and $Y^-$ denote the pools of correct and incorrect trajectories, respectively. $\rho_{S}$ indicates selection based on HES ranking. And half of the positive and negative samples are sampled in the full batch.}
    \label{tab:rl_sampling_strategies}
    \resizebox{\textwidth}{!}{
    \begin{tabular}{l c c l}
        \toprule
        \textbf{Strategy} & \textbf{Positive Samples ($P$)} & \textbf{Negative Samples ($N$)} & \textbf{Rationale} \\
        \midrule
        \rowcolor{gray!15} \textbf{Pos-High, Neg-Rand (Ours)} & $\text{Top-}\rho_{S}(Y^+)$ & $\text{Random}(Y^-)$ & Promotes complexity mastery \& error diversity \\
        Pos-Rand, Neg-Rand & $\text{Random}(Y^+)$ & $\text{Random}(Y^-)$ & Unbiased stochastic baseline \\
        Pos-High, Neg-Low & $\text{Top-}\rho_{S}(Y^+)$ & $\text{Bottom-}\rho_{S}(Y^-)$ & Maximizes positive-negative contrast \\
        Pos-Rand, Neg-Low & $\text{Random}(Y^+)$ & $\text{Bottom-}\rho_{S}(Y^-)$ & Evaluates impact of weak negatives \\
        Pos-Length/Difficulty, Neg-Rand & Top-Length / Difficulty & $\text{Random}(Y^-)$ & Benchmarks against heuristic criteria \\
        \bottomrule
    \end{tabular}
    }
    \vspace{-15pt}
\end{table*}

%% file: section/5_experiment_0128.tex
\input{table_0128/sft_main}
\input{table_0128/sft_proxy}
\input{table_0128/sft_detail}

\input{table_0128/sft_domain}

\section{Experiments}
\subsection{Experimental Setup}
\label{sec:exp_setup}

To ensure a fair and consistent evaluation, we define the common baselines and experimental configurations used across the SFT, RFT, and RL stages.

\paragraph{Baselines.} We compare our proposed method against the following selection strategies:
\textbf{Random} uniformly samples a subset of data from the available pool, serving as the standard baseline.
\textbf{Length} prioritizes samples with the longest token counts.
\textbf{Difficulty} selects samples with intermediate difficulty scores, filtering out both trivial and extremely hard cases.
\textbf{Highest-HES} selects the top samples ranked by our HES metric, as defined in Section~\ref{sec:method_selection}.
\textbf{Lowest-HES} selects the bottom samples ranked by HES, serving as a negative control to validate the sample quality captured by our metric.

\paragraph{Datasets and Models.}
We conduct experiments across three domains: Math, STEM, and Code. Detailed dataset statistics and training hyperparameters are provided in Appendix~\ref{sec:training_details}.
For \textbf{SFT}, our primary experiments are conducted using Qwen3-8B-Base~\citep{yang2025qwen3} on the \textit{Mixture-of-Thoughts} dataset~\citep{openr1}.\footnote{SFT results for Qwen3-4B-Base and DeepSeek-R1-Distill-Qwen-7B, and RFT experiments using DeepSeek-R1-Distill-Llama-8B are provided in Appendix~\ref{sec:supp_results}. And because the Code and STEM subsets of Mixture-of-Thoughts do not provide difficulty metrics, there is no difficulty baseline.}
For \textbf{RFT}, experiments are performed using DeepSeek-R1-Distill-Qwen-7B~\citep{deepseekai2025} on the \textit{DeepScaleR} dataset~\citep{deepscaler2025}. 
For \textbf{RL}, we also utilize the \textit{DeepScaleR} dataset~\citep{deepscaler2025}. Due to the high computational cost of RL, we adopt the smaller DeepSeek-R1-Distill-Qwen-1.5B~\citep{deepseekai2025}.

\paragraph{Evaluation Settings.}
We evaluate model performance on widely-recognized benchmarks, including AIME24~\citep{aops2024aime,aops2024aime2}, AIME25~\citep{aops2025aime,aops2025aime2}, HMMT23~\citep{hmmt2023}, HMMT24~\citep{hmmt2024}, HMMT25~\citep{hmmt2025}, OlymMATH~\citep{sun2025challenging}, GPQA-Diamond~\citep{rein2024gpqa}, and LiveCodeBench~\citep{jain2025livecodebench}. We use a temperature of 0.6 and a maximum generation length of 32,768. 

\subsection{SFT Experiments}
\label{sec:sft_main_results}

\paragraph{Main Results.} 
To evaluate the efficacy of HES as a unified metric, we investigate its performance across Math, Code, and STEM domains. 
We compare HES against random and heuristic baselines in two settings: the \textit{Efficiency Test} and \textit{De-noising Test}.
As shown in Table~\ref{tab:sft_0128_main} and~\ref{tab:sft_0128_code}, in the \textit{Efficiency Test}, HES consistently outperforms all baselines, even surpassing the full dataset in STEM, which indicates effective filtration of redundant signals. In the \textit{De-noising Test}, pruning the bottom 20\% of HES-ranked data yields the best overall performance in Math and Code, significantly beating the full dataset baseline. Meanwhile, the lowest-HES 20\% subset performs drastically worse than random selection, confirming that the metric accurately isolates harmful training noise.

\paragraph{Analysis on Efficiency and Transferability.}
To assess the efficiency of HES for large-scale data curation, we investigate whether small, computationally efficient models can serve as effective proxies for selecting data for larger foundation models.
We utilize a lightweight 0.6B model to screen the training data and then train the larger 8B model on the selected subset, comparing the results against the 8B model's self-selection.
As shown in Table~\ref{tab:sft_0128_proxy}, the performance achieved using the 0.6B proxy model is comparable to that of the 8B model's self-selection (30.20\% vs. 31.34\%), reducing the inference cost for data selection by over an order of magnitude without compromising data quality. This strong cross-model consistency suggests that HES captures the intrinsic reasoning complexity inherent to the data rather than model-specific artifacts, confirming its value as a cost-effective metric for massive-scale pipelines.

\label{sec:sft_analysis}

\paragraph{Comparable Analysis with Other Entropy Metrics.}
We compare HES against alternative entropy-based metrics in Table~\ref{tab:sft_0128_entropy}. HES consistently outperforms all alternatives: averaging metrics (AvgE, AvgHE) underperform by diluting sparse reasoning signals with abundant low-entropy tokens, and while ES is competitive, HES achieves superior results by strictly targeting the top percentile of uncertainty. Moreover, $\text{HES}_{\textit{absolute}}$ also underperforms the relative variant as its fixed threshold cannot adapt to the varying scales across different samples, confirming the importance of the adaptive, percentile-based formulation. These results validate that the cumulative magnitude of critical forking points, captured adaptively, is the most effective proxy for reasoning quality.

\begin{figure*}[!t]
    \centering
    \includegraphics[width=\linewidth]{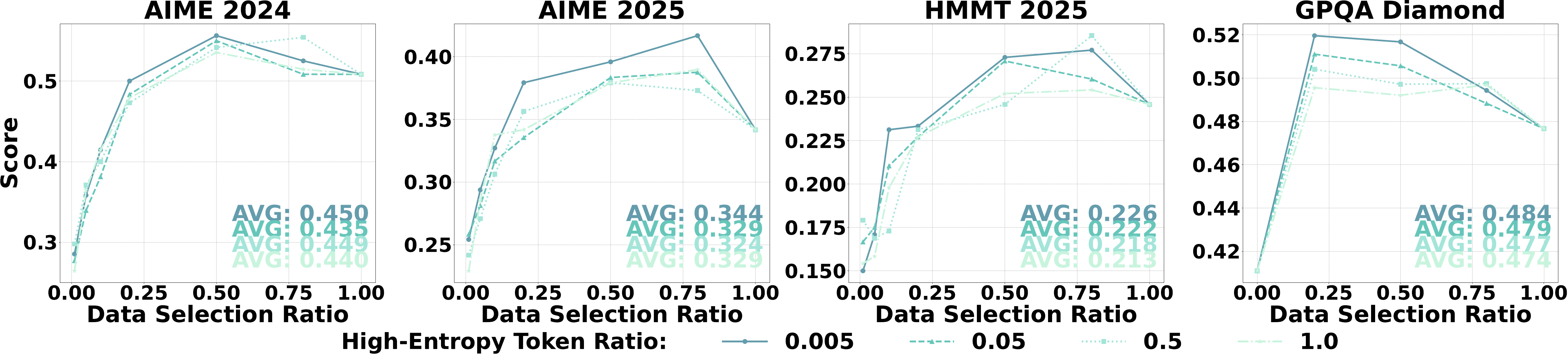}
    \vspace{-2em}
    \caption{Parameter sensitivity across diverse domains, including data selection ratio and high-entropy token ratio.}
    \label{args}
\end{figure*}

\input{table_0128/rft}

\paragraph{Sensitivity Analysis.}
To validate HES's robustness, we analyze its sensitivity to two key hyperparameters in Figure \ref{args}: data selection ratio and high-entropy token ratio across diverse domains. We find peak performance is almost consistently achieved when using approximately 20\% or 80\% of the data as ranked by HES, confirming that pruning the lowest-quality samples is a reliable strategy for improvement. While HES is robust to the specific high-entropy token ratio used, smaller, more targeted ratio of 0.005 delivers the best performance.

\input{table_0128/rl}

\subsection{RFT Experiments}

\paragraph{Main Results.}

We evaluate the robustness of HES across different selection scopes in RFT by investigating both localized \textit{Per-Query} and broader \textit{Global Pool Selection}, as detailed in Table~\ref{tab:rft_0128}. HES demonstrates consistent superiority over random baselines in both settings, proving its robustness regardless of the selection scope. Notably, the \textit{Per-Query Selection} often yields higher performance than the corresponding \textit{Global Pool Selection}. We attribute this advantage to the structural guarantee of query coverage in the \textit{Per-Query Selection}, which suggests that while HES filters for intrinsic quality, maintaining data diversity remains equally essential for maximizing model generalization.

\paragraph{Analysis on Scalability.}
We scale the candidate pool size ($k$) from 2 to 8 to examine HES in sample-rich environments. HES consistently outperforms random selection, achieving Per-Query gains of +1.18, +1.56, and +0.80 points for $k=2, 4, 8$ respectively. The stable efficacy across expanding search spaces confirms HES as a reliable signal for identifying high-quality trajectories regardless of candidate volume.

\subsection{RL Experiments}
\paragraph{Main Results.}
We evaluate an asymmetric sampling strategy that pairs Highest-HES positive samples with randomly selected negative ones, against baselines defined in Table~\ref{tab:rl_sampling_strategies}. As shown in Table~\ref{tab:rl_0128}, \textit{Pos-High, Neg-Rand} achieves the best average accuracy of 19.26\%, significantly outperforming random sampling (17.51\%) and even surpassing the Full-Batch baseline (18.34\%) with only 50\% of the data per step. This confirms that prioritizing high-quality positives is a more potent training signal than using larger, undifferentiated batches.

\paragraph{Analysis on Necessity of Negative Diversity.}
Complementing our positive selection strategy, we examine whether curating the negative sample pool yields benefits similar to prioritizing high-quality positives.
We compare strategies that constrain negative samples to specific quality tiers (e.g., Neg-Low) against the random negative sampling approach. Strategies that constrain the negative pool consistently degrade performance, falling even below the random baseline, indicating that restricting negative samples to the lowest-quality tier proves counterproductive. In contrast, random negative sampling proves essential for exposing the model to a diverse, unbiased distribution of failure modes, which is required to build robust generalization.

%% file: table_0128/sft_main.tex
\begin{table*}[htbp]
\centering
\setlength{\tabcolsep}{3.5pt}
\caption{Main results of SFT performance on \textbf{Math} domain. Results are reported as average pass@1 accuracy over 16 sampling paths (Average@16 \%). \textbf{Bold} indicates the best performance per benchmark.}
\label{tab:sft_0128_main}
\scalebox{0.9}{%
\begin{tabular}{@{}llcccccccc@{}}
\toprule
\textbf{Method} & \textbf{Ratio} & \textbf{AIME24} & \textbf{AIME25} & \textbf{HMMT23} & \textbf{HMMT24} & \textbf{HMMT25} & \textbf{Oly(E)} & \textbf{Oly(H)} & \textbf{AVG} \\
\midrule
Fullset & 100 & 50.83 & 34.17 & 35.21 & 28.13 & 24.58 & 42.94 & \textbf{6.94} & 31.83 \\
Random & 20 & 39.79 & 27.92 & 24.79 & 20.21 & 20.42 & 30.00 & 3.88 & 23.86 \\
Length & 20 & 46.67 & 34.79 & 31.04 & 24.79 & 23.54 & 36.06 & 6.44 & 29.05 \\
Difficulty & 20 & 38.13 & 27.50 & 25.83 & 22.50 & 18.96 & 26.50 & 3.69 & 23.30 \\
Lowest-HES & 20 & 18.54 & 18.96 & 11.88 & 11.04 & 7.92 & 11.00 & 2.94 & 11.75 \\
\rowcolor{gray!15} Highest-HES & 20 & 50.00 & 37.92 & 35.00 & 26.46 & 23.33 & 41.00 & 5.69 & 31.34 \\
\rowcolor{gray!15} Highest-HES & 80 & \textbf{52.50} & \textbf{41.67} & \textbf{37.50} & \textbf{29.17} & \textbf{27.71} & \textbf{47.00} & 6.69 & \textbf{34.61} \\
\bottomrule
\end{tabular}%
}
\end{table*}

%% file: table_0128/sft_proxy.tex
\begin{table*}[!t]
\centering
\setlength{\tabcolsep}{3.5pt}
\caption{Analysis of small-to-large model transferability. We compare the performance of the Qwen3-8B model trained on data selected by proxy models of varying sizes (0.6B) versus self-selection (8B). \textbf{Bold} indicates the best.}
\label{tab:sft_0128_proxy}
\scalebox{0.88}{
\begin{tabular}{@{}llcccccccc@{}}
\toprule
\textbf{Method} & \textbf{Ratio} & \textbf{AIME24} & \textbf{AIME25} & \textbf{HMMT23} & \textbf{HMMT24} & \textbf{HMMT25} & \textbf{Oly(E)} & \textbf{Oly(H)} & \textbf{AVG} \\
\midrule
\rowcolor{gray!15} Highest-HES (8B) & 20 & \textbf{50.00} & \textbf{37.92} & \textbf{35.00} & \textbf{26.46} & \textbf{23.33} & 41.00 & 5.69 & \textbf{31.34} \\
\rowcolor{gray!15} Highest-HES (0.6B) & 20 & 49.17 & 35.42 & 32.50 & 24.17 & 22.29 & \textbf{41.50} & \textbf{6.38} & 30.20 \\
\bottomrule
\end{tabular}
}
\end{table*}

%% file: table_0128/sft_detail.tex
\begin{figure*}[!t]
    \centering
    \begin{minipage}{\linewidth}
        \centering
        \setlength{\tabcolsep}{3.5pt}
        \captionof{table}{Ablation study on different entropy-based metrics for SFT data selection. \textbf{Bold} indicates the best.}
        \label{tab:sft_0128_entropy}
        \scalebox{0.88}{
            \begin{tabular}{@{}llcccccccc@{}}
            \toprule
            \textbf{Method} & \textbf{Ratio} & \textbf{AIME24} & \textbf{AIME25} & \textbf{HMMT23} & \textbf{HMMT24} & \textbf{HMMT25} & \textbf{Oly(E)} & \textbf{Oly(H)} & \textbf{AVG} \\
            \midrule
            Highest-AvgE & 20 & 40.21 & 31.25 & 28.54 & 23.13 & 19.17 & 31.19 & 4.88 & 25.48 \\
            Highest-AvgHE & 20 & 42.29 & 29.79 & 31.46 & 22.08 & 18.75 & 33.06 & 4.31 & 25.96 \\
            Highest-ES & 20 & 47.92 & 34.17 & 30.42 & 25.63 & 22.71 & 38.50 & 5.25 & 29.23 \\
            Highest-HES\textsubscript{\textit{absolute}} & 20 & 42.29 & 32.29 & 32.92 & \textbf{29.38} & 19.38 & 36.06 & \textbf{6.94} & 28.47 \\
            \rowcolor{gray!15} Highest-HES & 20 & \textbf{50.00} & \textbf{37.92} & \textbf{35.00} & 26.46 & \textbf{23.33} & \textbf{41.00} & 5.69 & \textbf{31.34} \\
            \bottomrule
            \end{tabular}
        }
    \end{minipage}
    
\vspace{-0.5em} 
\end{figure*}

%% file: table_0128/sft_domain.tex
\begin{table}[htbp]
    \centering
    \setlength{\tabcolsep}{4pt}
    \caption{SFT performance on \textbf{STEM} and \textbf{Code} domains. GPQA results are reported as Average@16, LiveCodeBench results are reported as Average@10 following \citet{jain2025livecodebench}. \textbf{Bold} indicates the best.}
    \label{tab:sft_0128_code}
    \label{tab:sft_0128_stem}
    \scalebox{0.9}{
    \begin{tabular}{@{}llcc@{}}
        \toprule
        \textbf{Method} & \textbf{Ratio} & \textbf{GPQA} & \textbf{LiveCodeBench} \\
        \midrule
        Fullset & 100 & 47.66 & 58.76 \\
        Random & 20 & 45.86 & 49.06 \\
        Length & 20 & 50.82 & 53.84 \\
        Lowest-HES & 20 & 31.44 & 35.32 \\
        \rowcolor{gray!15} Highest-HES & 20 & \textbf{51.96} & 54.43 \\
        \rowcolor{gray!15} Highest-HES & 80 & 49.43 & \textbf{61.65} \\
        \bottomrule
    \end{tabular}
    }
    \vspace{-10pt}
\end{table}

%% file: table_0128/rft.tex
\begin{table*}[!t]
\centering
\small
\caption{Main results of RFT experiments across different candidate pool sizes ($k$). \textbf{Bold} indicates the best.}
\label{tab:rft_0128}

\scalebox{0.98}{%
\begin{tabular}{@{}lcccccccc@{}}
\toprule
\textbf{Method} & \textbf{AIME24} & \textbf{AIME25} & \textbf{HMMT23} & \textbf{HMMT24} & \textbf{HMMT25} & \textbf{Oly(E)} & \textbf{Oly(H)} & \textbf{AVG} \\
\midrule
\multicolumn{9}{c}{\textit{Per-Query Selection ($k=2$)}} \\
\cdashline{1-9} 
Random & 46.46 & 34.17 & 31.88 & \textbf{28.54} & 21.46 & 35.63 & 4.31 & 28.92 \\
Length & 46.04 & 33.33 & 33.75 & 28.13 & 19.58 & 35.94 & 4.81 & 28.80 \\
Lowest-HES & 43.75 & 34.17 & 30.63 & 25.00 & 18.96 & 33.81 & 4.44 & 27.25 \\
\rowcolor{gray!15} Highest-HES & \textbf{48.33} & \textbf{34.58} & \textbf{34.38} & 28.13 & \textbf{21.67} & \textbf{38.06} & \textbf{5.56} & \textbf{30.10} \\
\midrule
\multicolumn{9}{c}{\textit{Global Pool Selection ($k=2$)}} \\
\cdashline{1-9} 
Random & 42.29 & 31.04 & 30.00 & \textbf{26.88} & 18.54 & 30.81 & 4.75 & 26.33 \\
Length & 45.21 & \textbf{35.42} & 27.92 & 25.21 & 18.33 & 32.13 & 5.13 & 27.05 \\
Difficulty & 45.00 & 32.71 & 28.13 & 23.54 & 17.71 & 30.13 & 4.31 & 25.93 \\
Lowest-HES & 19.38 & 14.58 & 10.21 & 11.04 & 7.92 & 10.81 & 2.00 & 10.85 \\
\rowcolor{gray!15} Highest-HES & \textbf{46.46} & 34.58 & \textbf{32.50} & 25.42 & \textbf{20.83} & \textbf{34.00} & \textbf{5.31} & \textbf{28.44} \\
\midrule
\multicolumn{9}{c}{\textit{Per-Query Selection ($k=4$)}} \\
\cdashline{1-9} 
Random & 48.33 & 31.88 & 33.33 & 27.71 & 18.75 & 36.00 & 4.50 & 28.64 \\
Length & 46.04 & 33.33 & \textbf{33.75} & 28.13 & 19.58 & 35.94 & 4.81 & 28.80 \\
Lowest-HES & 43.75 & 33.54 & 30.42 & 26.46 & \textbf{21.67} & 32.69 & 4.75 & 27.61 \\
\rowcolor{gray!15} Highest-HES & \textbf{48.96} & \textbf{36.25} & \textbf{33.75} & \textbf{29.17} & 21.04 & \textbf{37.13} & \textbf{5.13} & \textbf{30.20} \\
\midrule
\multicolumn{9}{c}{\textit{Global Pool Selection ($k=4$)}} \\
\cdashline{1-9} 
Random & 45.42 & \textbf{38.13} & \textbf{30.83} & \textbf{26.04} & 20.21 & 33.06 & 4.63 & 28.33 \\
Length & 45.21 & 35.42 & 27.92 & 25.21 & 18.33 & 32.13 & 5.13 & 27.05 \\
Difficulty & 45.00 & 32.71 & 28.12 & 23.54 & 17.71 & 30.13 & 4.31 & 25.93 \\
Lowest-HES & 21.46 & 18.96 & 12.92 & 11.04 & 9.38 & 11.56 & 1.69 & 12.43 \\
\rowcolor{gray!15} Highest-HES & \textbf{48.96} & 33.13 & 30.00 & 24.38 & \textbf{21.88} & \textbf{36.19} & \textbf{5.56} & \textbf{28.59} \\
\midrule
\multicolumn{9}{c}{\textit{Per-Query Selection ($k=8$)}} \\
\cdashline{1-9} 
Random & 49.17 & 34.58 & 29.38 & 28.75 & \textbf{22.08} & \textbf{36.69} & 4.69 & 29.33 \\
Length & 46.67 & 35.42 & 32.29 & \textbf{30.21} & 20.63 & 35.94 & 5.63 & 29.54 \\
Lowest-HES & 42.50 & 32.50 & \textbf{32.71} & 28.54 & 20.00 & 34.69 & 4.88 & 27.97 \\
\rowcolor{gray!15} Highest-HES & \textbf{50.42} & \textbf{37.92} & 30.42 & 28.96 & \textbf{22.08} & 35.31 & \textbf{5.81} & \textbf{30.13} \\
\midrule
\multicolumn{9}{c}{\textit{Global Pool Selection ($k=8$)}} \\
\cdashline{1-9} 
Random & 46.67 & 35.42 & 31.67 & 28.33 & 18.54 & 33.06 & 4.19 & 28.27 \\
Length & 48.75 & 36.04 & 30.42 & 26.46 & \textbf{22.08} & 36.44 & \textbf{5.56} & 29.39 \\
Difficulty & 44.38 & 32.29 & 31.04 & 24.79 & 20.42 & 31.50 & 4.63 & 27.01 \\
Lowest-HES & 25.63 & 21.25 & 18.96 & 14.66 & 11.67 & 14.50 & 2.44 & 15.59 \\
\rowcolor{gray!15} Highest-HES & \textbf{48.96} & \textbf{36.46} & \textbf{31.88} & \textbf{29.17} & 19.38 & \textbf{38.31} & 5.19 & \textbf{29.91} \\
\bottomrule
\end{tabular}%
}
\vspace{-10pt}
\end{table*}

%% file: table_0128/rl.tex

\begin{table*}[htbp]
\centering
\caption{RL results comparing different sampling strategies. \textbf{Bold} indicates the best performance per benchmark.}
\label{tab:rl_0128}
\setlength{\tabcolsep}{1pt}
\scalebox{0.92}{
\begin{tabular}{@{}lcccccccc@{}}
\toprule
\textbf{Method} & \textbf{AIME24} & \textbf{AIME25} & \textbf{HMMT23} & \textbf{HMMT24} & \textbf{HMMT25} & \textbf{Oly(E)} & \textbf{Oly(H)} & \textbf{AVG} \\
\midrule
Full-Batch & 33.33 & 25.63 & 17.30 & 14.00 & \textbf{15.21} & 19.69 & 3.19 & 18.34 \\
Pos-Rand, Neg-Rand & 32.30 & 25.21 & 13.54 & 14.11 & 13.96 & 19.18 & 4.25 & 17.51 \\
Pos-Rand, Neg-Low & 33.75 & 24.79 & 15.42 & 13.73 & 11.67 & 19.00 & 3.38 & 17.39 \\
Pos-Low, Neg-Rand & 30.42 & 24.17 & 15.63 & 13.10 & 14.17 & 19.50 & 3.25 & 17.18 \\
Pos-High, Neg-Low & 31.25 & 25.42 & 15.21 & 13.96 & 12.71 & 18.88 & 3.56 & 17.28 \\
Pos-Length, Neg-Rand & 31.04 & 26.25 & 16.46 & 13.95 & 14.79 & \textbf{20.00} & 4.06 & 18.08 \\
Pos-Difficulty, Neg-Rand & 35.00 & 24.79 & 16.88 & 13.77 & 14.17 & 18.94 & 3.75 & 18.19 \\
\rowcolor{gray!15} Pos-High, Neg-Rand & \textbf{35.42} & \textbf{27.29} & \textbf{17.92} & \textbf{18.13} & 11.88 & 19.88 & \textbf{4.31} & \textbf{19.26} \\
\bottomrule
\end{tabular}
}
\vspace{-5pt}
\end{table*}

%% file: section/6_related_work.tex
\section{Related Work}

\paragraph{Training-Time Efficiency.}
Improving training-time efficiency is a critical challenge across the dominant paradigms for LLM reasoning: SFT, RFT, and RL. To address this, some works focus on curating higher-quality training sets through methods like verification-based selection or preference optimization~\citep{hu2022lora,rein2024gpqa,yang2025qwen3,DBLP:journals/corr/abs-2308-01825}. Other works aim to provide more granular learning signals by directly supervising intermediate steps with process-level rewards~\citep{lightman2023let,cui2025process,luo2025ursa}. While powerful, the effectiveness of these methods hinges on a reliable signal to identify which reasoning paths or steps are valuable, and they often rely on costly external reward models or human annotations. Our work addresses this need by introducing HES, a simple, intrinsic metric that provides this critical signal uniformly across all three paradigms.
\vspace{-5pt}
\paragraph{Data Selection.}
Data selection aims to identify optimal training subsets from large-scale datasets~\citep{albalak2024survey,li2024datacomp}. Traditional approaches rely on heuristic rules~\citep{robertson2009probabilistic,xie2023data} or global metrics like perplexity~\citep{wettig2024qurating,marion2023less}. While gradient-based methods offer finer-grained selection~\citep{killamsetty2021grad,han2023context}, their computational costs scale prohibitively with sequence length and model size. Alternative approaches using task-specific models~\citep{shum2025predictive} or LLM-based selection~\citep{toshniwal2025genselect} introduce substantial overhead.
Current data selection thus faces two challenges: (1) the inability to identify crucial tokens through global metrics, and (2) the lack of efficient metrics deployable across different training paradigms. Our HES addresses both limitations.

%% file: section/7_conclusion.tex
\section{Conclusion}
In this paper, we introduced \textbf{High-Entropy Sum (HES)}, a novel, training-free metric designed to efficiently identify high-quality reasoning data.
We proposed a unified data selection framework based on HES and validated its effectiveness and efficiencyacross three dominant training paradigms: SFT, RFT, and RL.
In conclusion, our work established HES as a powerful, cost-effective solution for scaling up LLM reasoning capabilities, highlighting the pivotal role of critical tokens in data selection.

\section*{Limitations}
While HES proves effective across SFT, RFT, and RL, several limitations remain.
First, although we validate HES on Math, Code, and STEM domains, its applicability to other reasoning-intensive tasks such as agentic planning has not been explored.
Second, HES is designed for long-CoT reasoning where critical forking points are sparse; for tasks with uniformly distributed uncertainty (e.g., open-ended generation), the metric's discriminative power may diminish.
Finally, our current framework treats HES as a static score computed once before training; integrating HES as a dynamic, curriculum-based signal that adapts during training remains an open direction.

\section*{Ethical Considerations}
This work focuses on improving the training efficiency of LLMs through data selection and does not involve the collection or use of private, sensitive, or personally identifiable data. All datasets used in our experiments are publicly available and have been widely adopted by the research community. Our method is designed to enhance reasoning quality by filtering low-quality training data, which may also help reduce the risk of models learning from flawed or misleading reasoning paths. We do not foresee direct negative societal impacts from this work, though we acknowledge that improvements in LLM reasoning capabilities should be accompanied by responsible deployment practices.

%% file: section/appendix.tex
\appendix
\section{Usage of LLMs}
During the writing process, we used LLMs to assist with language polishing and proofreading. All LLM-assisted outputs were carefully reviewed and verified by the human authors. The authors take full responsibility for the entire content of this paper, including the accuracy of all claims and the final wording of the text.

\section{Broader Impact}
This paper presents a unified, computation-efficient framework for data selection in LLM reasoning. Our work has several broader impacts on the community:

\paragraph{Environmental Sustainability.} 
The massive computational cost of training LLMs poses significant environmental challenges. By demonstrating that training on a small, rigorously curated subset (e.g., 20\%) of data can match or even exceed the performance of full-dataset training, our method directly contributes to reducing the energy consumption and carbon footprint associated with model post-training. Furthermore, unlike methods relying on training expensive Reward Models, our proposed HES metric is training-free and computationally lightweight, further minimizing the resource overhead of the data curation pipeline itself.

\paragraph{Democratization of Research.} 
A key finding of our work is the effective small-to-large transferability, where small models (e.g., 0.6B) can serve as accurate proxies to select data for large foundation models. This capability significantly lowers the barrier to entry for academic laboratories and smaller organizations with limited computing resources, empowering a broader community to contribute to high-quality dataset construction and model development without requiring massive GPU clusters.

\paragraph{Model Reliability.} 
By effectively identifying and removing low-quality data—which often contains hallucinations, inconsistent logic, or noise—our method contributes to building more reliable and robust reasoning systems. Reducing a model's exposure to flawed reasoning paths during training helps mitigate the risk of low-quality reasoning, where models generate plausible-sounding but factually or logically incorrect outputs.

\section{Detailed Related Work}
\paragraph{Efficient Reasoning.}
Recent progress on efficient reasoning has advanced along two directions: (1) test-time efficiency~\citep{jaech2024openai,snell2024scaling} and (2) training-time efficiency. In the former, mainstream methods broaden the explored trajectory set by deepening CoT and branching into multiple paths~\citep{wei2022chain,sessa2024bond,yao2023tree,besta2024graph,sel2023algorithm,ning2023skeleton,fu2025deep}. Building on this, some work incorporates tool use, heuristic control, or self-reflection to verify processes or outcomes and stabilize search~\citep{chen2022program,gou2023tora,yu2025chain,renze2024self}. 
However, these methods often require substantial token budgets.
This motivates efforts to compress outputs by extracting key information from reasoning trajectories~\citep{zhang2025lightthinker,xu2025chain,aytes2025sketch}.
On the training-time side, the focus is on efficiently converting exploration-derived signals into policy updates.
Dominant pipelines improve reasoning by verification-based selection or preference optimization~\citep{hu2022lora,rein2024gpqa,yang2025qwen3,DBLP:journals/corr/abs-2308-01825}.
In addition, some work explicitly targets process quality by supervising intermediate steps with process-level signals~\citep{lightman2023let,cui2025process,luo2025ursa}.
Therefore, efficient reasoning either allocates more computation to critical branches at test time or concentrates training budgets on samples or spans with higher pedagogical value at training time. 
Both require a stable signal that reliably identifies critical fork tokens. Our work is mainly focused on training-time efficiency, and we leave the exploration of test-time efficiency to future work.

\paragraph{Data Selection.}
Data selection takes the full training data as input and chooses a subset to train~\citep{albalak2024survey}.
Large-scale, high-quality datasets play a decisive role in training effectiveness~\citep{li2024datacomp,DBLP:conf/nips/PenedoKALMRW024}.
Traditional selection methods rely on rules and heuristics, which make it difficult to assess the quality and complexity of reasoning processes~\citep{robertson2009probabilistic,xie2023data}.
Therefore, many works adopt global proxies such as perplexity, average token entropy, sample length, or heuristic difficulty~\citep{wettig2024qurating,marion2023less,ivison2022data}. 
Yet these metrics inevitably mix routine low-entropy spans with the sparse high-entropy tokens that signal critical decision points, which dilutes the signal and makes it hard to distinguish correct solutions with low process complexity from those with high process complexity.
More fine-grained strategies rely on selection criteria derived from gradients or influence estimates~\citep{killamsetty2021grad,han2023context,xia2024less}.
However, their compute and memory costs scale rapidly with sequence length and model size, which limits their practicality for large-scale training.
Consequently, current data selection faces two core gaps: (1) the lack of a model-intrinsic signal that directly captures process-level critical decision points; (2) the lack of a metric that can be deployed consistently and efficiently across diverse training methods without introducing new supervision. 
Our proposed HES addresses both gaps.

\section{Training Details}
\label{sec:training_details}
\paragraph{SFT Training Details.}
Our implementation is based on the \texttt{open-r1}~\citep{openr1} framework, and we follow the recommended SFT hyperparameters. We use the AdamW optimizer with a learning rate of $4\times 10^{-5}$. The learning rate follows a cosine decay schedule with a warm-up ratio of 0.1. All experiments are run with a global batch size of 64 and are trained for a total of 3 epochs.

\textbf{RFT Training Details.}
First, we sample 32 candidate responses using the model for each query in the dataset. Then, we filter these candidates to retain only the correct trajectories. This creates a pool which serves as the foundation for selection. Training settings, such as the optimizer and learning rate schedule, remain consistent with SFT, with the only difference being the curated training dataset.

\textbf{RL Training Details.}
Our implementation is built upon the \texttt{verl} codebase~\citep{sheng2025hybridflow} and follows the standard GRPO settings. Consistent with the DeepScaleR-1.5B-Preview~\citep{deepscaler2025} setup, the maximum generation length is set to 8192 tokens. We use a temperature of 0.6 and generate 32 rollouts per query. We train for 3 epochs, a total of 628 steps. It is worth noting that we train the baseline to its officially reported accuracy.

\section{Datasets}
\label{sec:dataset}

\paragraph{Mixture-of-Thoughts~\citep{openr1}} is a comprehensive compilation of high-quality chain-of-thought reasoning data, specifically assembled to enhance model capabilities across diverse logic-intensive domains, including Math, Code and STEM domain. This multi-domain dataset is aggregated from several rigorously selected sources. It features a strong foundation in mathematical reasoning, incorporating the challenging OpenR1-Math and Open-Math-Reasoning collections. To ensure cross-domain generalization and robustness, the collection is significantly broadened with specialized subsets from other fields, including complex programming tasks from the Codeforces-CoT dataset and scientific reasoning problems from the STEM subset of the Llama-Nemotron dataset.

\paragraph{DeepScaler~\citep{deepscaler2025}}  is a curated collection of approximately 40,000 unique mathematics problem-answer pairs, specifically assembled to train advanced problem-solving models. This comprehensive dataset is compiled from a variety of high-quality and challenging sources. It features a deep historical archive of problems from prestigious competitions, including the American Invitational Mathematics Examination (AIME) from 1984 to 2023 and the American Mathematics Competition (AMC) from years prior to 2023. To ensure breadth and diversity, the collection is further supplemented with content from the Omni-MATH and Still datasets.

\paragraph{OpenR1-Math-220k~\citep{openr1}} is a massive dataset designed to advance mathematical reasoning in LLMs. It comprises 220,000 math problems sourced from the NuminaMath 1.5 collection.
A key feature of this dataset is that each problem is accompanied by two to four distinct reasoning traces, or step-by-step solutions, generated by the DeepSeek R1 model. To ensure reliability, these traces underwent a rigorous verification process: most were validated using Math Verify, while 12\% were assessed by a Llama-3.3-70B-Instruct judge. Critically, every problem in the dataset is guaranteed to contain at least one reasoning trace that leads to a correct answer.
The dataset is organized into two distinct splits: (1) The default split, containing 94,000 problems, is the recommended version for training. It has been shown to yield the best performance improvements after SFT.
(2) The extended split expands the collection to 131,000 problems by incorporating additional data sources. While this provides a greater volume of reasoning traces, models fine-tuned on this split have shown slightly lower performance, which is likely because the questions from the added sources are less difficult than those in the core dataset.

\paragraph{Open-Math-Reasoning~\citep{moshkov2025aimo}} is also a large-scale dataset created to train LLMs in advanced mathematical reasoning. At its core, it contains 306,000 unique mathematical problems sourced from the Art of Problem Solving (AoPS) forums. These are complemented by a massive volume of solutions, including 3.2 million long-CoT examples and 1.7 million Tool-Integrated Reasoning (TIR) solutions, showcasing diverse and detailed problem-solving paths. The dataset also includes 566,000 ``GenSelect'' samples, which are designed to teach models how to evaluate and select the most promising solution from multiple candidates. To further expand its utility, an additional 193,000 problems from the AoPS forums are included without solutions. The creation process involved state-of-the-art models, with Qwen2.5-32B-Instruct used for preprocessing problems and both DeepSeek-R1 and QwQ-32B for generating the high-quality solutions. Notably, this dataset was a foundational component of the winning submission to the AIMO-2 Kaggle competition, highlighting its power and effectiveness in building top-tier AI reasoning systems.

To maintain a balanced distribution and manageable computational cost across all domains, we constructed the \textit{Full-Dataset} by randomly sampling approximately 100,000 examples from the source collections of each domain (Math, Code, and STEM). This standardized subset serves as the baseline for all comparative experiments.

\section{Computational Resources}
\label{sec:computational_resources}
All experiments were conducted on a single node equipped with 8$\times$ NVIDIA A100 80GB GPUs, 224 CPU cores, and 2048 GB of system memory. The total computational cost of all experiments reported in this paper amounts to approximately 500 GPU hours.

\section{Licenses}

\paragraph{Datasets:} OpenR1-Math-220k~\citep{openr1} is released under the Apache 2.0 License. Open-Math-Reasoning~\citep{openr1} is released under the CC-BY-4.0 License. DeepScaleR~\citep{deepscaler2025} is released under the MIT License. Mixture-of-Thoughts~\citep{openr1} is released under the Apache 2.0 License.
\paragraph{Models:} Qwen3-8B-Base~\citep{yang2025qwen3}, Qwen3-4B-Base~\citep{yang2025qwen3}, and Qwen3-0.6B~\citep{yang2025qwen3} are released under the Apache 2.0 License. DeepSeek-R1-Distill-Qwen-7B~\citep{deepseekai2025} and DeepSeek-R1-Distill-Qwen-1.5B~\citep{deepseekai2025} are released under the MIT License. DeepSeek-R1-Distill-Llama-8B~\citep{deepseekai2025} is released under the MIT License and additionally governed by the Llama 3.1 Community License Agreement.

All datasets and models are used in this work for their originally intended purposes, i.e., training and evaluating LLMs on reasoning tasks, which is fully consistent with the terms of their respective licenses.

\input{table/sft_2}
\input{table/sft_3}

\input{table/sft_5}
\input{table/rft_4}

\input{table_0128/sft_length}

\section{Supplementary Results}
\label{sec:supp_results}

To further validate the robustness and generalizability of the HES metric, we conducted additional experiments across different datasets, model sizes, and architectures. This section presents detailed results that complement the main findings.

\subsection{Consistency Across Datasets and Model Scales}
We replicated the SFT experiments on the OpenR1-Math-220k dataset using the Qwen3-8B-Base model (Table~\ref{tab:sft_results_updated}) and on the Open-Math-Reasoning dataset using the smaller Qwen3-4B-Base model (Table~\ref{tab:sft_results_new}).

The results consistently demonstrate the efficacy of HES-based data selection:
\begin{itemize}
    \item \textbf{Superiority of Highest-HES:} In both settings, training on the top 20\% of data ranked by HES (Highest-HES) achieves performance comparable to or better than random selection, often approaching the full-dataset baseline.
    \item \textbf{Effective Noise Filtering:} Training on the top 80\% of data (i.e., pruning the bottom 20\% with the lowest HES) consistently yields the best overall performance. For instance, on OpenR1-Math-220k, the Highest-HES (80) strategy achieves an average accuracy of 31.13\%, surpassing the full-dataset performance of 29.38\%. Similarly, on Open-Math-Reasoning with the 4B model, the top 80\% subset achieves an average of 29.14\%, outperforming the full dataset's 26.63\%.
    \item \textbf{Detrimental Effect of Low-HES Data:} Consistent with our main findings, training on the bottom 20\% of data ranked by HES (Lowest-HES) leads to a catastrophic drop in performance, significantly underperforming even the random baseline. This confirms that HES effectively identifies low-quality data that can be harmful to model training.
\end{itemize}

\subsection{Robustness Across Model Architectures}
To ensure that the effectiveness of HES is not limited to the Qwen model family, we conducted RFT experiments using the DeepSeek-R1-Distilled-Llama-8B model (Table~\ref{tab:llama_rft_results}). We compared random selection, HES-based selection, and inverse HES selection across both per-query and global pool settings with $k=4$.

The results on the Llama-based model align with our previous findings:
\begin{itemize}
    \item \textbf{HES Outperforms Random Selection:} In both per-query and global pool settings, selecting the highest HES responses consistently yields better average performance than random selection. For example, in the per-query setting, Highest-HES achieves an average of 13.23\%, compared to 12.42\% for random selection.
    \item \textbf{Criticality of Global Pool Selection:} The discriminative power of HES is particularly evident in the global pool setting. While random selection achieves 11.85\%, Highest-HES improves this to 13.54\%. Conversely, selecting the lowest HES responses results in a severe performance degradation to 4.24\%, highlighting the metric's ability to distinguish high-quality reasoning paths from poor ones across different model architectures.
\end{itemize}

\subsection{Decoupling Quality from Length}
To ensure that HES captures true reasoning density rather than merely acting as a proxy for length, we analyze its performance independent of token count.
We stratify the Math subset into Long, Medium, and Short groups and compare the performance of the top 20\% (Highest-HES) versus the bottom 20\% (Lowest-HES) within each length interval, as shown in Table~\ref{tab:sft_0128_length}. Highest-HES consistently outperforms Lowest-HES across all length groups, confirming that the metric functions independently of sample length. The discrimination power is most pronounced in the Short group, where Highest-HES (15.57\%) nearly triples the accuracy of Lowest-HES (5.50\%), indicating that identifying critical steps remains crucial even for shorter reasoning chains. In the Long group, Highest-HES maintains a distinct advantage despite the inherent complexity of longer samples, proving that HES effectively captures thought density rather than simply favoring longer sequences.

These supplementary results reinforce the conclusion that HES is a robust, model-agnostic metric for data selection in LLM reasoning, capable of improving performance across various datasets, model scales, and architectures.

\input{figure/distribution}
\input{figure/wordcloud}

\section{Analysis}
\subsection{Discriminative Ability of HES}
To validate the effectiveness of HES as a quality metric, we analyze its ability to discriminate between high- and low-quality samples against several alternative entropy-based metrics. This analysis is conducted across multiple benchmarks, including AIME24, HMMT23, HMMT24, and HMMT25.
As illustrated by the score distributions in Figure \ref{aime_2024} to Figure \ref{hmmt_2025}, the HES demonstrates a significantly clearer separation between correct and incorrect reasoning paths compared to other metrics. Specifically, the distributions of HES scores for correct and incorrect samples show minimal overlap. The mean HES for incorrect paths is substantially higher than for correct paths, indicating that flawed reasoning corresponds to a higher cumulative sum of uncertainty at pivotal moments.
In contrast, global metrics like the AveE and even the more targeted AvgHET show less separation between their means for correct and incorrect samples. This analysis confirms that the cumulative signal captured by HES is a more robust and reliable indicator of reasoning quality than metrics based on either global averages or the average intensity of pivotal moments alone. The clear separation validates its use as a foundational metric for our data selection framework.

Additionally, Figure \ref{aime_2025_2} provides specific entropy value references.

\subsection{High-entropy Token Distribution}
To better understand the characteristics of our training data, we conduct a detailed analysis of the token-level entropy distributions for both the OpenR1-Math-220k and Open-Math-Reasoning datasets. The analysis, illustrated in Figure \ref{distribution1} and \ref{distribution2}, reveals that the entropy distribution is heavily right-skewed. 
The vast majority of tokens exhibit very low entropy, corresponding to routine or predictable steps, while a small minority of tokens accounts for the high-entropy tail of the distribution. 
Our calculations show that the entropy value corresponding to the top 0.5\% of high-entropy tokens is approximately 1.55 for OpenR1-Math-220k and 1.60 for Open-Math-Reasoning. This data-driven finding provides an empirical basis for our threshold choice of $1.6$ in the $HES_{absolute}$ experiments. 
Furthermore, a qualitative analysis using word clouds shown in Figure \ref{fig:wordclouds_combined} confirms that these high-entropy tokens are not random noise; they typically correspond to semantically significant forking points that guide the reasoning process. Taken together, these findings validate our core premise: a small, identifiable subset of tokens represents the most complex decision points, motivating our focus on the HES metric.

\subsection{Token Distribution Example}
The entropy distribution of a sample is shown as follows. We can see that high entropy words are small in number of forking points, while low entropy words are large in number.

\section{Computational Efficiency and Cost Analysis}
\label{sec:computational_cost}

A primary concern in data selection is whether the cost of computing the metric outweighs the benefits of reduced training data. In this section, we analyze the computational overhead of HES in both online generation (RFT/RL) and offline screening (SFT) scenarios, demonstrating that HES is a highly cost-effective metric.

\subsection{Online Settings (RFT and RL): Zero Extra Inference Overhead}
In RFT and RL, the model generates reasoning trajectories (rollouts) in real-time. In these scenarios, calculating HES incurs \textbf{zero extra inference overhead} in terms of model forward passes.

\begin{itemize}
    \item \textbf{Piggybacking on Generation:} During the auto-regressive generation process, the model must perform a forward pass at every time step $t$ to compute the probability distribution over the vocabulary, $P(x_t | x_{<t})$, in order to sample the next token.
    \item \textbf{Synchronous Calculation:} The entropy calculation is performed synchronously using these intermediate logits. We simply calculate the entropy of the sampled token's distribution and accumulate it in memory. This operation involves only negligible vector operations (complexity $O(V)$, where $V$ is vocabulary size), which is orders of magnitude faster than the matrix multiplications required for the model's forward pass.
    \item \textbf{No Additional Forward Pass:} Unlike methods that require a separate reward model or a second pass over the completed sequence, HES reuses the computation already performed for generation. Thus, it does not increase the inference latency or compute budget during rollout.
\end{itemize}

\subsection{Offline Settings (SFT): Efficient Screening via Proxy Models}
For SFT, data selection is typically performed offline on static datasets. While calculating HES requires a forward pass to obtain logits, we demonstrate that this cost can be minimized by using smaller proxy models.

\paragraph{Small Model Transferability:} We investigated whether HES scores derived from smaller models correlate with data quality for larger models. As shown in Table~\ref{tab:sft_0128_proxy}, we used a Qwen3-0.6B model to screen data for training a Qwen3-8B model.
\begin{itemize}
    \item \textbf{Performance Consistency:} The 8B model trained on data selected by the 0.6B proxy achieved an average accuracy of 30.20\%, comparable to the performance when using the 8B model for selection (31.34\%).
    \item \textbf{Cost Reduction:} The inference cost of a 0.6B model is approximately $1/13$th that of an 8B model. By using a small proxy model, we can screen massive datasets at a fraction of the cost.
\end{itemize}

\subsection{Overall Return on Investment (ROI)}
The computational cost of HES selection should be weighed against the savings in training compute.
\begin{itemize}
    \item \textbf{Training vs. Inference:} The computational cost of training (backward pass + optimizer) is significantly higher than inference (forward pass only).
    \item \textbf{Data Pruning:} Our experiments show that training on the top 20\% of HES-selected data matches or exceeds full-dataset performance. This represents an \textbf{80\% reduction} in total training FLOPs.
\end{itemize}

In summary, the inference cost of calculating HES (especially when using small proxy models) is negligible compared to gradient updates on 80\% of low-value data, which makes HES a highly efficient strategy.

\input{figure/aime_2025_2}
\input{figure/aime_2024}
\input{figure/hmmt_2023}
\input{figure/hmmt_2024}
\input{figure/hmmt_2025}

\includepdf[pages=-,scale=0.95]{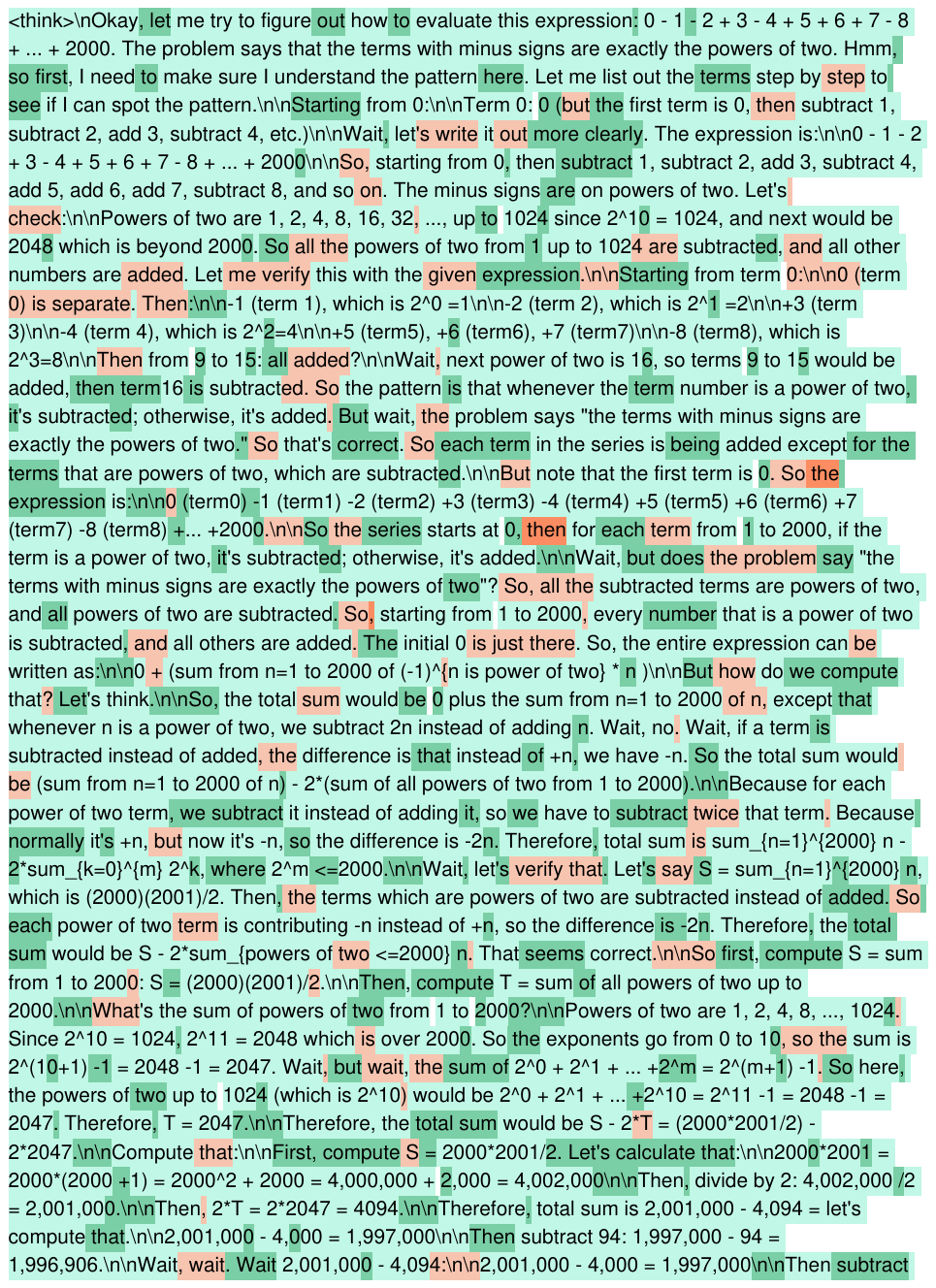}

%% file: table/sft_2.tex
\begin{table*}[t]
\centering
\setlength{\tabcolsep}{1pt}
\caption{Performance comparison of SFT using Qwen3-8B-Base on OpenR1-Math-220k.}
\label{tab:sft_results_updated}

\scalebox{0.93}{
\begin{tabular}{@{}llcccccccc@{}}
\toprule
\textbf{Method} & \textbf{Ratio} & \textbf{AIME24} & \textbf{AIME25} & \textbf{HMMT23} & \textbf{HMMT24} & \textbf{HMMT25} & \textbf{Oly(E)} & \textbf{Oly(H)} & \textbf{AVG} \\
\midrule
Fullset & 100 & \textbf{48.96} & 32.50 & 31.67 & 25.83 & 22.50 & 38.38 & 5.81 & 29.38 \\
Random & 20 & 30.00 & 30.21 & 24.17 & 19.79 & 17.50 & 25.00 & 5.56 & 21.75 \\
Highest-AvgE & 20 & 39.38 & 26.67 & 22.29 & 19.38 & 18.75 & 26.38 & 4.38 & 22.46 \\
Highest-AvgHE & 20 & 39.58 & 29.58 & 23.54 & 18.75 & 20.00 & 29.13 & 3.63 & 23.46 \\
Lowest-HES & 20 & 15.42 & 12.29 & 6.04 & 8.13 & 6.04 & 5.63 & 2.13 & 7.95 \\
\rowcolor{gray!15} Highest-HES & 20 & 48.54 & 32.92 & 30.63 & 23.33 & 22.29 & 37.94 & \textbf{6.56} & 28.89 \\
\rowcolor{gray!15} Highest-HES & 80 & \textbf{48.96} & \textbf{35.63} & \textbf{36.04} & \textbf{26.67} & \textbf{22.71} & \textbf{41.88} & 6.00 & \textbf{31.13} \\
\bottomrule
\end{tabular}
}
\end{table*}

%% file: table/sft_3.tex
\begin{table*}[t]
\centering
\setlength{\tabcolsep}{1pt}
\caption{Performance comparison of SFT using Qwen3-4B-Base on Open-Math-Reasoning.}
\label{tab:sft_results_new}

\scalebox{0.93}{
\begin{tabular}{@{}llcccccccc@{}}
\toprule
\textbf{Method} & \textbf{Ratio} & \textbf{AIME24} & \textbf{AIME25} & \textbf{HMMT23} & \textbf{HMMT24} & \textbf{HMMT25} & \textbf{Oly(E)} & \textbf{Oly(H)} & \textbf{AVG} \\
\midrule
Fullset & 100 & 39.17 & 31.04 & 29.58 & 24.17 & 20.83 & 37.00 & 4.63 & 26.63 \\
Random & 20 & 30.83 & 26.04 & 20.42 & 16.67 & 17.08 & 20.06 & 3.25 & 19.19 \\
Highest-AvgE & 20 & 34.58 & 28.13 & 21.46 & 14.79 & 17.92 & 23.63 & 3.06 & 20.51 \\
Highest-AvgHE & 20 & 33.75 & 25.42 & 21.67 & 17.29 & 15.83 & 23.56 & 2.19 & 19.96 \\
Lowest-HES & 20 & 16.04 & 16.46 & 9.17 & 7.08 & 6.46 & 7.00 & 2.13 & 9.19 \\
\rowcolor{gray!15} Highest-HES & 20 & 38.13 & 31.88 & 22.71 & 20.42 & 18.54 & 28.00 & \textbf{4.88} & 23.51 \\
\rowcolor{gray!15} Highest-HES & 80 & \textbf{46.25} & \textbf{33.96} & \textbf{31.88} & \textbf{25.83} & \textbf{21.46} & \textbf{39.44} & \textbf{5.13} & \textbf{29.14} \\
\bottomrule
\end{tabular}
}
\end{table*}

%% file: table/sft_5.tex
\begin{table*}[t]
\centering
\setlength{\tabcolsep}{1pt}
\caption{Performance comparison of SFT using DeepSeek-R1-Distilled-Qwen-7B on OpenR1-Math-220k. \textbf{Bold} indicates the best performance per benchmark.}
\label{tab:ds_sft_results}

\scalebox{0.95}{
\begin{tabular}{@{}llcccccccc@{}}
\toprule
\textbf{Method} & \textbf{Ratio} & \textbf{AIME24} & \textbf{AIME25} & \textbf{HMMT23} & \textbf{HMMT24} & \textbf{HMMT25} & \textbf{Oly(E)} & \textbf{Oly(H)} & \textbf{AVG} \\
\midrule
Full-Dataset & 100 & 46.25 & 33.13 & 31.67 & 26.25 & 22.29 & 38.50 & 5.50 & 29.08 \\
Random & 20 & 42.08 & 37.08 & 29.79 & 26.67 & 23.13 & 38.13 & 5.13 & 28.86 \\
Lowest-HES & 20 & 29.38 & 27.29 & 17.50 & 18.33 & 14.58 & 19.25 & 3.06 & 18.48 \\
\rowcolor{gray!15} Highest-HES & 20 & \textbf{51.67} & \textbf{40.83} & 33.96 & \textbf{30.21} & 24.58 & \textbf{45.75} & \textbf{6.88} & \textbf{33.41} \\
\rowcolor{gray!15} Highest-HES & 80 & 50.42 & 35.21 & 35.21 & 26.25 & 24.17 & 42.25 & 5.44 & 31.28 \\
\bottomrule
\end{tabular}
}
\end{table*}

%% file: table/rft_4.tex
\begin{table*}[t]
\centering
\setlength{\tabcolsep}{2pt}
\caption{Performance comparison of RFT strategies using DeepSeek-R1-Distilled-Llama-8B.}
\label{tab:llama_rft_results}

\scalebox{0.9}{%
\begin{tabular}{@{}lcccccccc@{}}
\toprule
\textbf{Method} & \textbf{AIME24} & \textbf{AIME25} & \textbf{HMMT23} & \textbf{HMMT24} & \textbf{HMMT25} & \textbf{Oly(E)} & \textbf{Oly(H)} & \textbf{AVG} \\
\midrule
\multicolumn{9}{c}{\textit{Per-Query Selection ($k=4$)}} \\
\cdashline{1-9} 
Random & 17.08 & \textbf{22.50} & 10.63 & \textbf{12.71} & 9.17 & 12.38 & 2.44 & 12.42 \\
Lowest-HES & 14.58 & 18.13 & 12.08 & 10.21 & 9.38 & 12.06 & 2.06 & 11.21 \\
\rowcolor{gray!15} Highest-HES & \textbf{18.54} & 19.38 & \textbf{14.79} & 10.83 & \textbf{11.88} & \textbf{14.31} & \textbf{2.88} & \textbf{13.23} \\
\midrule
\multicolumn{9}{c}{\textit{Global Pool Selection ($k=4$)}} \\
\cdashline{1-9} 
Random & 14.17 & 19.58 & 11.04 & \textbf{11.46} & 10.56 & 12.81 & \textbf{3.35} & 11.85 \\
Lowest-HES & 1.46 & 10.00 & 5.21 & 5.21 & 2.92 & 3.56 & 1.31 & 4.24 \\
\rowcolor{gray!15} Highest-HES & \textbf{17.92} & \textbf{20.63} & \textbf{14.38} & 11.25 & \textbf{12.29} & \textbf{15.19} & 3.13 & \textbf{13.54} \\
\bottomrule
\end{tabular}%
}
\end{table*}

%% file: table_0128/sft_length.tex
\begin{table*}[h]
\centering
\setlength{\tabcolsep}{3.5pt}
\scalebox{0.9}{
\begin{tabular}{@{}llcccccccc@{}}
\toprule
\textbf{Length Group} & \textbf{Selection} & \textbf{AIME24} & \textbf{AIME25} & \textbf{HMMT23} & \textbf{HMMT24} & \textbf{HMMT25} & \textbf{Oly(E)} & \textbf{Oly(H)} & \textbf{AVG} \\
\midrule
\multirow{2}{*}{Long} & Lowest-HES & 35.00 & 30.21 & \textbf{31.88} & \textbf{22.92} & \textbf{20.83} & 28.88 & \textbf{4.63} & 24.91 \\
 & \cellcolor{gray!15} Highest-HES & \cellcolor{gray!15} \textbf{40.42} & \cellcolor{gray!15} \textbf{31.88} & \cellcolor{gray!15} 26.67 & \cellcolor{gray!15} 22.71 & \cellcolor{gray!15} 18.75 & \cellcolor{gray!15} \textbf{31.38} & \cellcolor{gray!15} 4.50 & \cellcolor{gray!15} \textbf{25.19} \\
\cdashline{1-10} 
\multirow{2}{*}{Medium} & Lowest-HES & 26.67 & 23.96 & 17.08 & 13.54 & 16.04 & 16.25 & 2.88 & 16.63 \\
 & \cellcolor{gray!15} Highest-HES & \cellcolor{gray!15} \textbf{40.63} & \cellcolor{gray!15} \textbf{29.17} & \cellcolor{gray!15} \textbf{25.63} & \cellcolor{gray!15} \textbf{23.75} & \cellcolor{gray!15} \textbf{19.17} & \cellcolor{gray!15} \textbf{28.44} & \cellcolor{gray!15} \textbf{3.38} & \cellcolor{gray!15} \textbf{24.31} \\
\cdashline{1-10} 
\multirow{2}{*}{Short} & Lowest-HES & 11.88 & 10.00 & 1.67 & 6.46 & 1.67 & 5.63 & 1.19 & 5.50 \\
 & \cellcolor{gray!15} Highest-HES & \cellcolor{gray!15} \textbf{28.13} & \cellcolor{gray!15} \textbf{20.42} & \cellcolor{gray!15} \textbf{17.08} & \cellcolor{gray!15} \textbf{12.08} & \cellcolor{gray!15} \textbf{12.50} & \cellcolor{gray!15} \textbf{16.75} & \cellcolor{gray!15} \textbf{2.00} & \cellcolor{gray!15} \textbf{15.57} \\
\bottomrule
\end{tabular}
}
\caption{Performance comparison of HES under different length groups. The Math subset is stratified into Long, Medium, and Short groups based on token length. Within each group, we compare the top 20\% (Highest-HES) versus the bottom 20\% (Lowest-HES) selections trained on Qwen3-8B. \textbf{Bold} indicates the better performance within each length group.}
\label{tab:sft_0128_length}
\end{table*}

%% file: figure/distribution.tex
\begin{figure}[ht!]   
    \centering

    \begin{subfigure}[b]{0.48\textwidth}
        \centering
        \includegraphics[width=\linewidth]{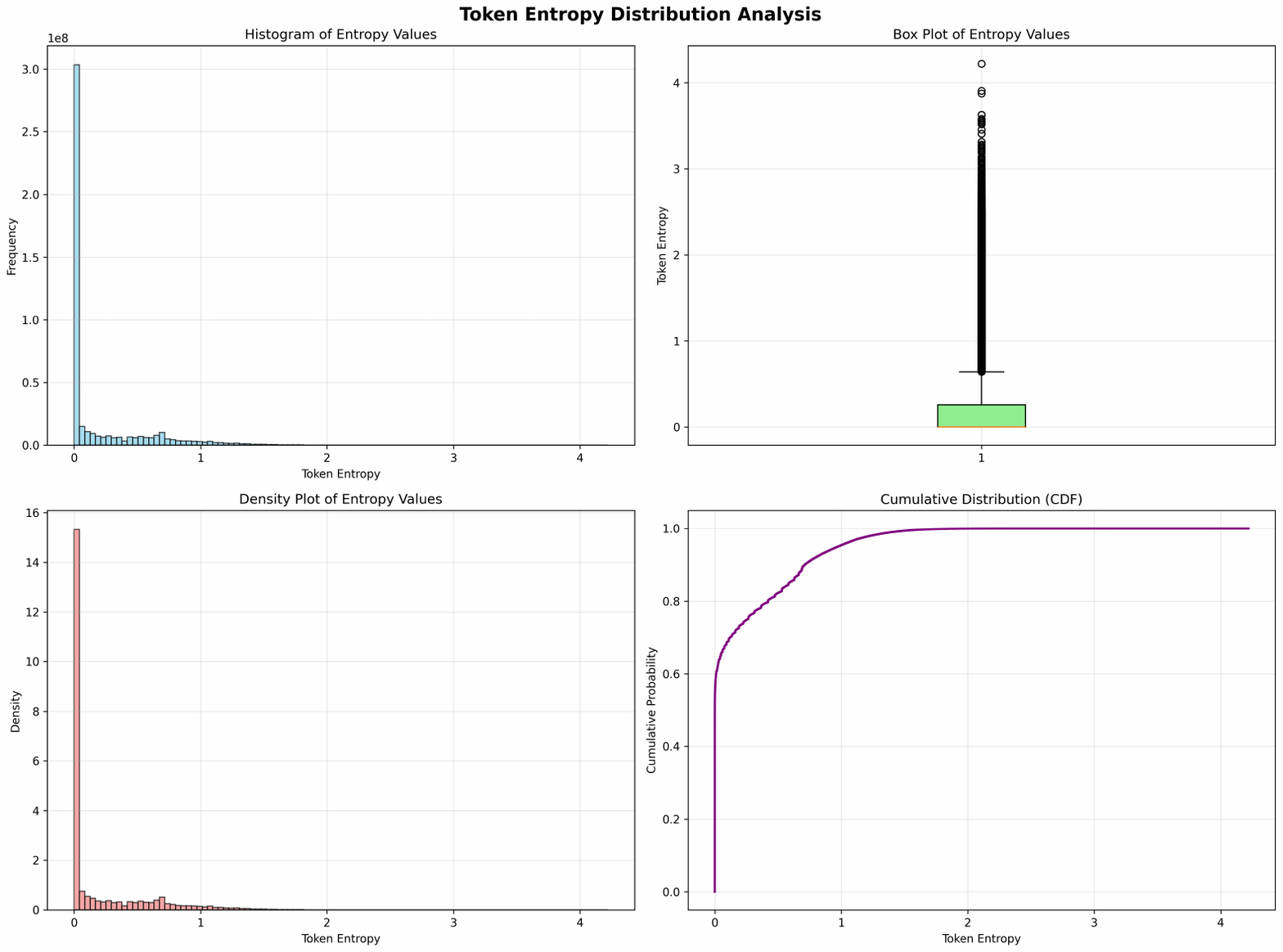}
        \caption{Token entropy distribution of OpenR1-Math-220k.}
        \label{distribution1} 
    \end{subfigure}
    \hfill 
    \begin{subfigure}[b]{0.48\textwidth}
        \centering
        \includegraphics[width=\linewidth]{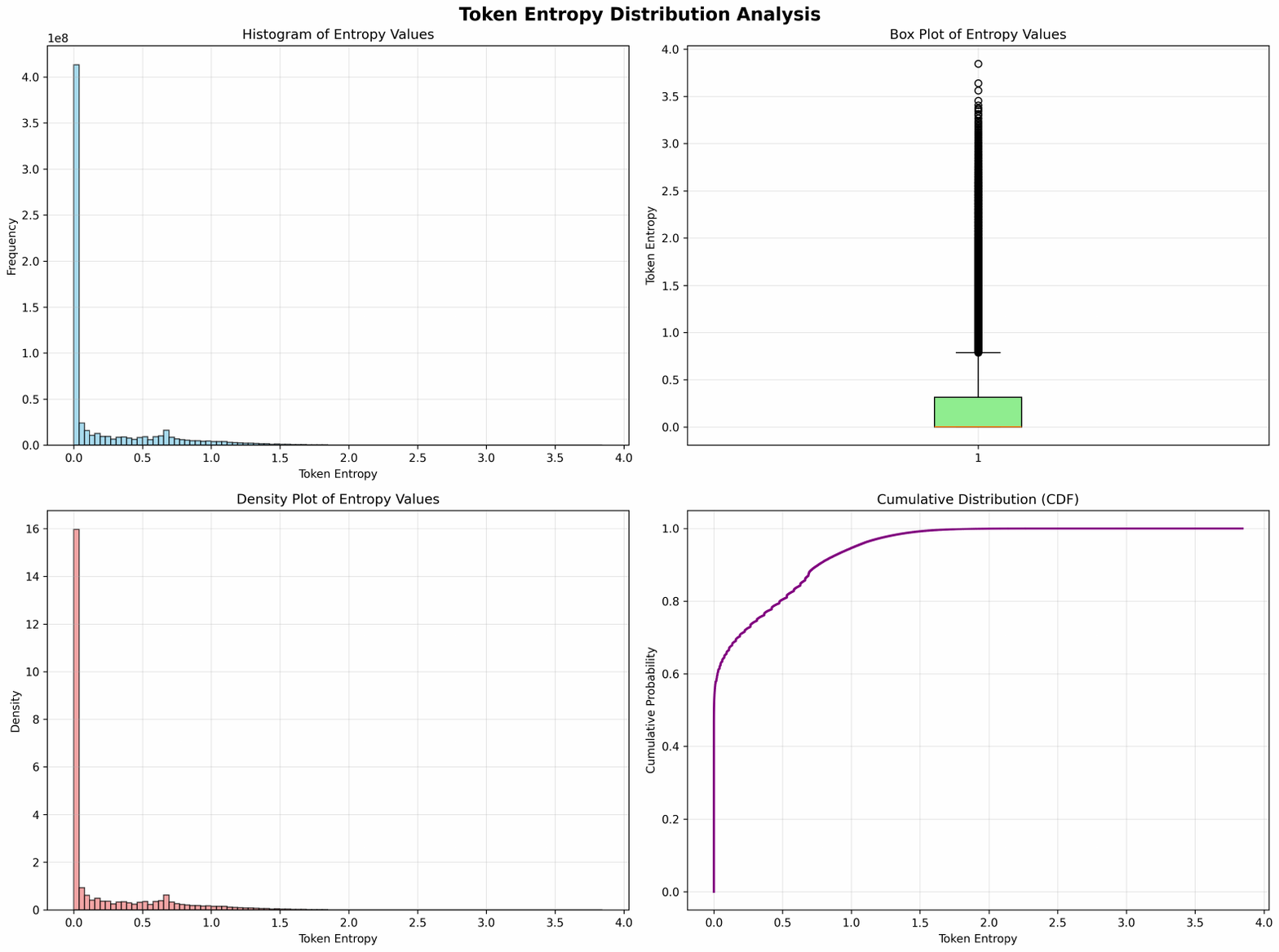}
        \caption{Token entropy distribution of Open-Math-Reasoning.}
        \label{fig:openmath} 
    \end{subfigure}
    \caption{Wordcloud of high-entropy tokens.}
    \label{distribution2} 
\end{figure}


%% file: figure/wordcloud.tex
\begin{figure}[ht!]   
    \centering

    \begin{subfigure}[b]{0.48\textwidth}
        \centering
        \includegraphics[width=\linewidth]{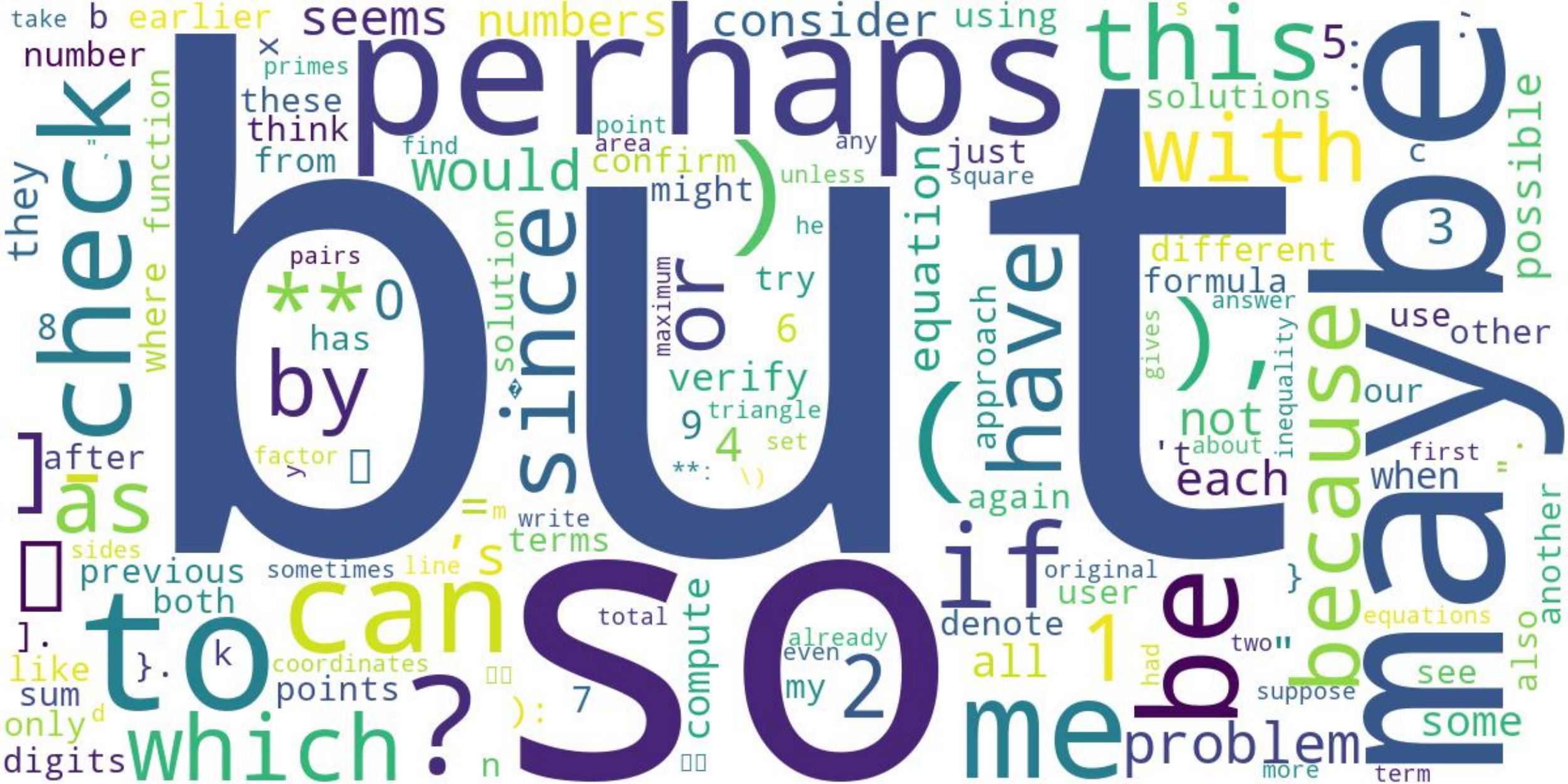}
        \caption{Wordcloud of OpenR1-Math-220k.}
        \label{fig:openr1} 
    \end{subfigure}
    \hfill 
    \begin{subfigure}[b]{0.48\textwidth}
        \centering
        \includegraphics[width=\linewidth]{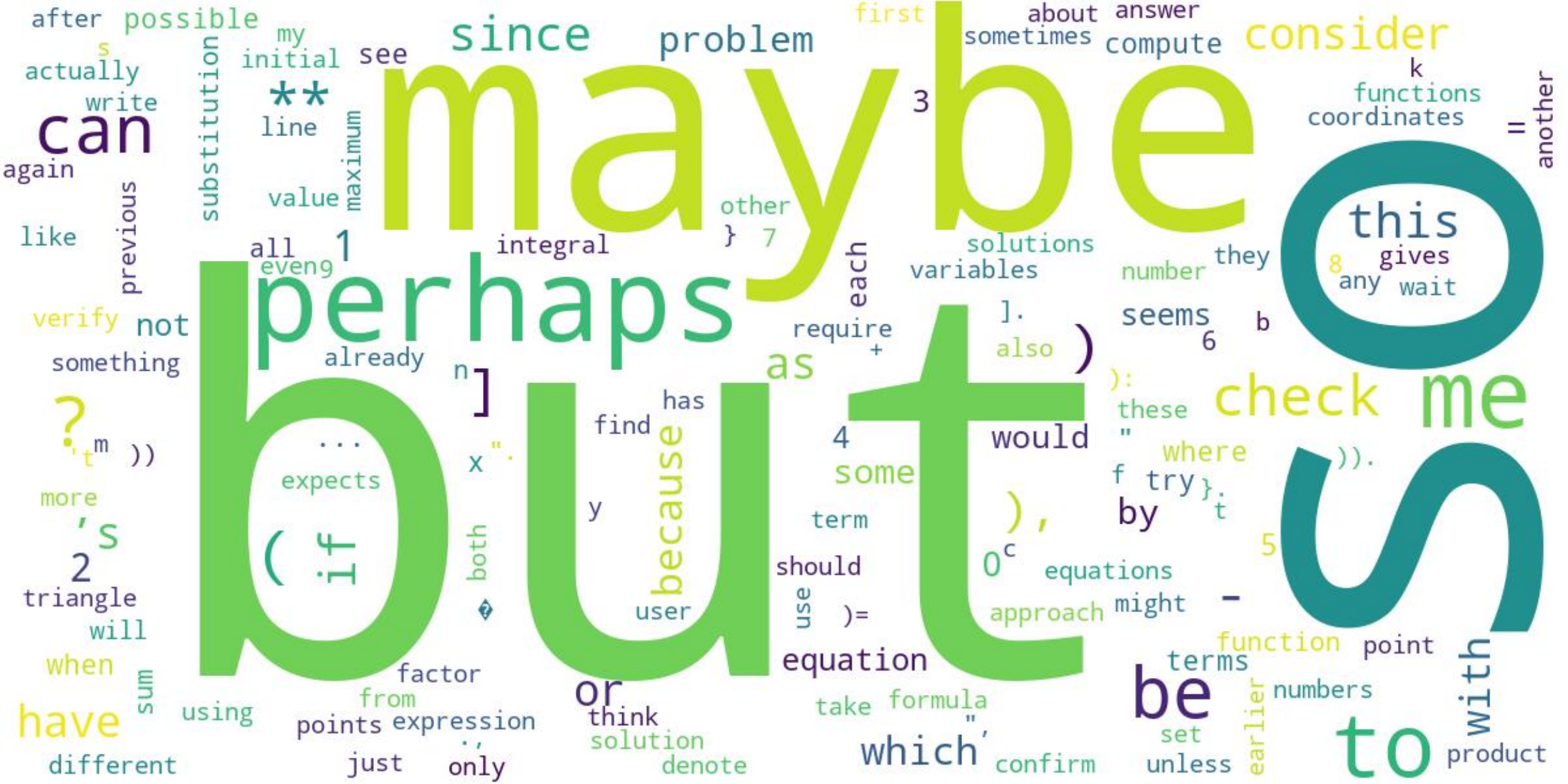}
        \caption{Wordcloud of Open-Math-Reasoning.}
        \label{fig:openmath} 
    \end{subfigure}
    \caption{Wordcloud of high-entropy tokens.}
    \label{fig:wordclouds_combined} 
\end{figure}

%% file: figure/aime_2025_2.tex
\begin{figure*}[t]   
\centering
\setlength{\abovecaptionskip}{-0.05cm}
\includegraphics[width=1.0\linewidth,scale=1.0]{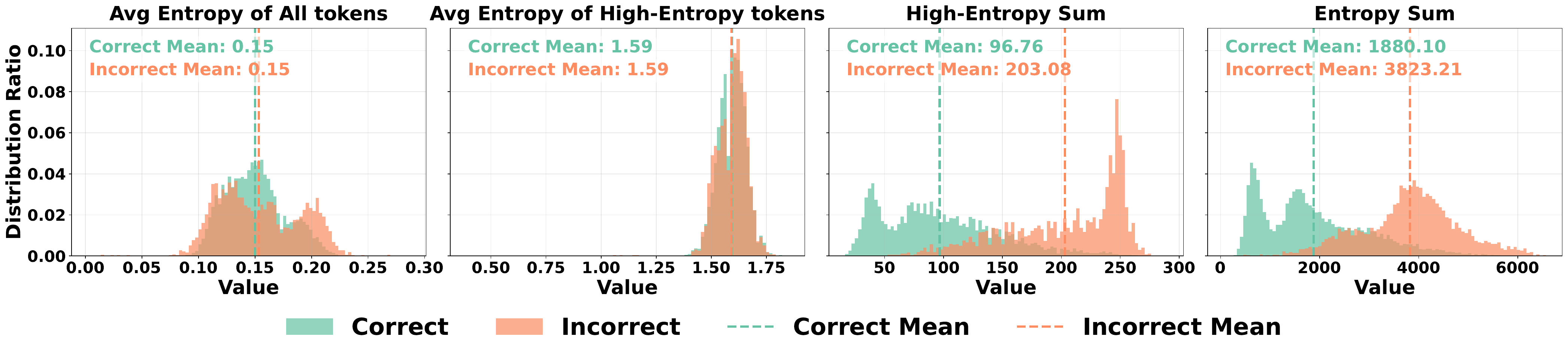}
\caption{Comparative analysis of discriminative ability between HES and other metrics based on 512 responses per problem sampled by Qwen3-14B on AIME 2025. The x-axis is specific entropy value.}
\label{aime_2025_2}
\end{figure*}

%% file: figure/aime_2024.tex
\begin{figure*}[t]   
\centering
\setlength{\abovecaptionskip}{-0.05cm}
\includegraphics[width=1.0\linewidth,scale=1.0]{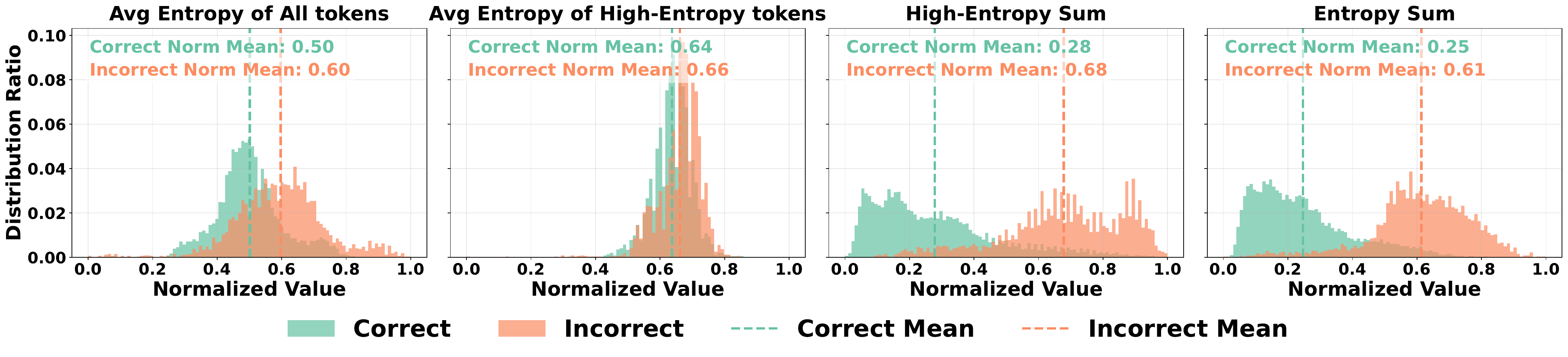}
\caption{Comparative analysis of discriminative ability between HES and other metrics based on 512 responses per problem sampled by Qwen3-14B on AIME 2024.}
\label{aime_2024}
\end{figure*}

%% file: figure/hmmt_2023.tex
\begin{figure*}[t]   
\centering
\setlength{\abovecaptionskip}{-0.05cm}
\includegraphics[width=1.0\linewidth,scale=1.0]{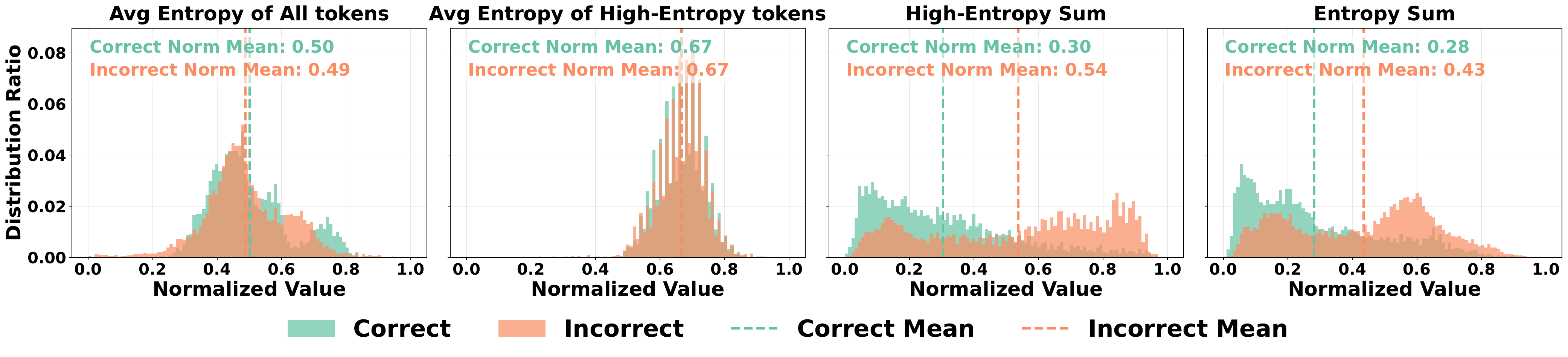}
\caption{Comparative analysis of discriminative ability between HES and other metrics based on 512 responses per problem sampled by Qwen3-14B on HMMT 2023.}
\label{hmmt_2023}
\end{figure*}

%% file: figure/hmmt_2024.tex
\begin{figure*}[t]   
\centering
\setlength{\abovecaptionskip}{-0.05cm}
\includegraphics[width=1.0\linewidth,scale=1.0]{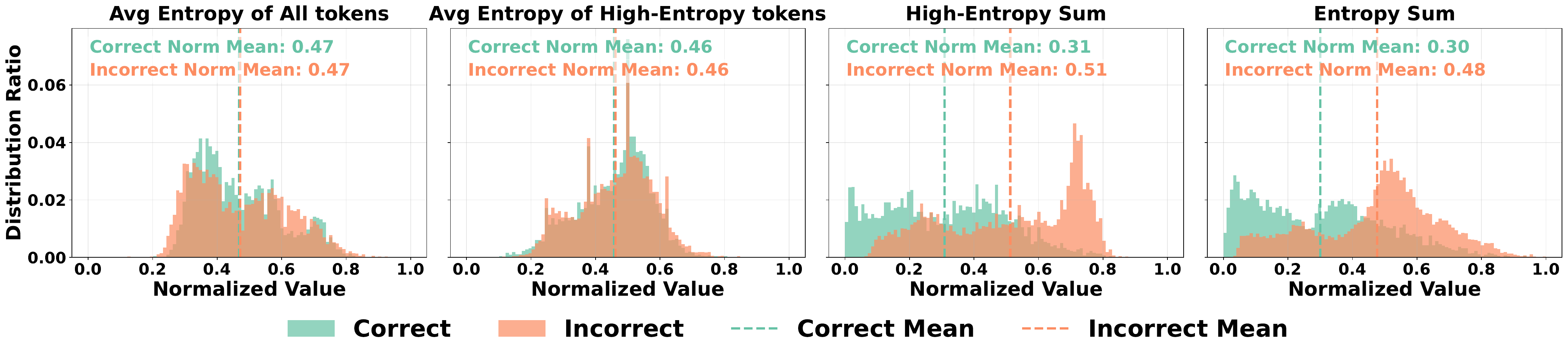}
\caption{Comparative analysis of discriminative ability between HES and other metrics based on 512 responses per problem sampled by Qwen3-14B on HMMT 2024.}
\label{hmmt_2024}
\end{figure*}

%% file: figure/hmmt_2025.tex
\begin{figure*}[t]   
\centering
\setlength{\abovecaptionskip}{-0.05cm}
\includegraphics[width=1.0\linewidth,scale=1.0]{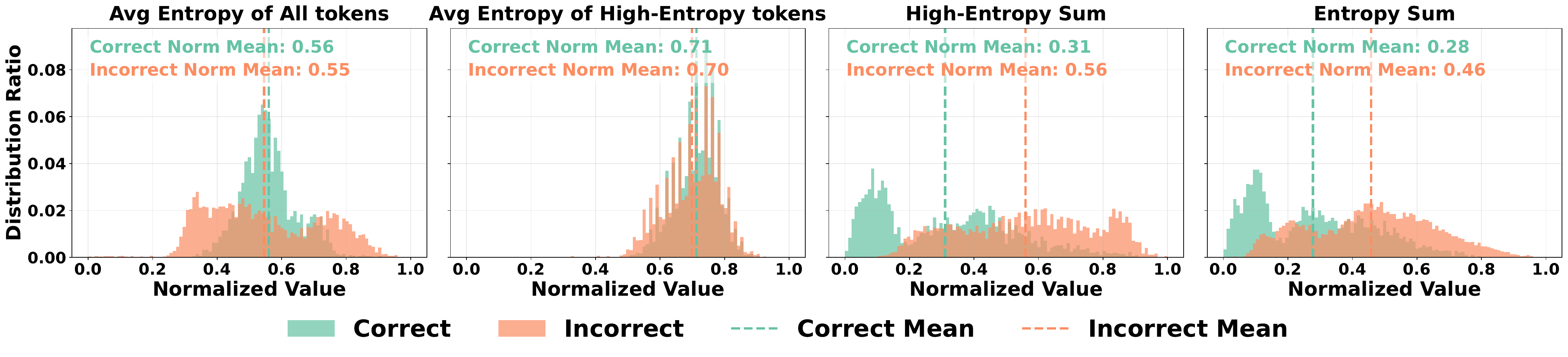}
\caption{Comparative analysis of discriminative ability between HES and other metrics based on 512 responses per problem sampled by Qwen3-14B on HMMT 2025.}
\label{hmmt_2025}
\end{figure*}